\documentclass[conf]{new-aiaa}
\usepackage[utf8]{inputenc}
\usepackage{graphicx}
\usepackage{amsmath}
\usepackage[version=4]{mhchem}
\usepackage{siunitx}
\usepackage{longtable}
\usepackage{placeins}
\usepackage{ulem}
\usepackage{caption}
\usepackage{xcolor}
\usepackage{colortbl}
\usepackage{pdflscape}
\usepackage{tabularx}
\usepackage[most]{tcolorbox}
\usepackage{enumitem}
\title{A framework for assuring the accuracy and fidelity of an AI-enabled Digital Twin of en route UK airspace}

\author{Adam Keane \footnote{Research Associate, Project Bluebird, The Alan Turing Institute}, Nick Pepper\footnote{Senior Research Associate, Project Bluebird, The Alan Turing Institute}, Chris Burr\footnote{Senior Researcher in Trustworthy Systems, The Alan Turing Institute}, Amy Hodgkin\footnote{Research Associate, Project Bluebird, The Alan Turing Institute}, and Dewi Gould\footnote{Research Associate, Project Bluebird, The Alan Turing Institute}}
\affil{The Alan Turing Institute, London, England, NW1 2DB, United Kingdom}
\author{John Korna\footnote{Senior Systems Engineer, Department of Research and Development, NATS} and Marc Thomas\footnote{Researcher, Department of Research and Development, NATS; also Visiting Professor, Queen Mary University London}}
\affil{NATS, Whiteley, Fareham, England, PO15 7FL, United Kingdom}

\begin{document}

\maketitle

\begin{abstract}

Digital Twins combine simulation, operational data and Artificial Intelligence (AI), and have the potential to bring significant benefits across the aviation industry. Project Bluebird, an industry-academic collaboration, has developed a probabilistic Digital Twin of en route UK airspace as an environment for training and testing AI Air Traffic Control (ATC) agents. Whilst the primary research focus is on enabling AI agent development, there are a number of potential applications of this technology within human Air Traffic Control Officer (ATCO) training and assessment, decision support tools, demand forecasting, conflict detection and resolution, and airspace redesign and optimisation. There is a developing regulatory landscape for this novel technology. Regulatory requirements are expected to be application specific, and may need to be tailored to each specific use case.

In this paper, we draw on emerging guidance for both Digital Twin development and the use of Artificial Intelligence/Machine Learning (AI/ML) in Air Traffic Management (ATM) to present an assurance framework. This framework defines actionable goals and the evidence required to demonstrate that a Digital Twin accurately represents its physical counterpart and also provides sufficient functionality across target use cases. It provides a structured approach for researchers to assess, understand and document the strengths and limitations of the Digital Twin, whilst also identifying areas where fidelity could be improved. Furthermore, it serves as a foundation for engagement with stakeholders and regulators, supporting discussions around the regulatory needs for future applications, and contributing to the emerging guidance through a concrete, working example of a Digital Twin.

The framework leverages a methodology known as Trustworthy and Ethical Assurance (TEA) to develop an assurance case. An assurance case is a nested set of structured arguments that provides justified evidence for how a top-level goal has been realised. We define our Goal Claim as: `the Digital Twin of en route UK airspace and Basic Training has sufficient accuracy and fidelity for its intended uses'. In this paper we provide an overview of each structured argument and a number of `deep dives' which elaborate in more detail upon particular arguments, including the required evidence,  assumptions and justifications.
\end{abstract}



\section{Introduction}
\label{sec:intro}

\lettrine{D}igital Twins have been defined in a variety of ways but a general consensus is that they are a virtual representation of a physical system, have predictive capability, and can be dynamically updated with data from the physical system \cite{WOOLEY2023940}. An AI-enabled or agentic Digital Twin is one which has the capacity to connect to and/or inform the decisions of AI agent(s). Given their data-driven nature, such systems lend themselves to the use or implementation of Artificial Intelligence and Machine Learning (AI/ML) \cite{singh}. Various guidelines have been released to create a best practice for their development, including domain-general frameworks and advice \cite{gemini_principles_2018, nasem2024}, and domain-specific guidance \cite{caa_cap2970}.

The use of automated technologies and Machine Learning in the field of Air Traffic Management (ATM) presents new opportunities, such as improved data analytics and decision support, while also posing novel challenges, particularly around security, safety, transparency, and explainability. However, many of these opportunities and risks remain poorly explored in the empirical literature \cite{nasem2024}.

During this period of uncertainty, regulatory bodies such as the Federal Aviation Administration, the UK Civil Aviation Authority and the European Union Aviation Safety Agency have responded by releasing draft guidance on how to assure key properties of such systems (e.g. safety, transparency, fairness) \cite{faa_roadmap, caa_ai_cap, easa_guidelines_2024}, which extend beyond traditional regulatory software requirements and guidance around verification and validation \cite{CAA_CAP670, EASA_ANS_CNS_AMC}, and include a greater emphasis on socio-technical factors, such as organisational readiness. At the time of writing, the UK Law Commission has an open commission examining the legal framework around autonomous flight, particularly concerning liability, to support its development \cite{law_commission}.

Project Bluebird is a partnership between NATS (the UK’s leading provider of air traffic management services), The Alan Turing Institute, and The University of Exeter \cite{bluebird}. Its core goal is to provide a training and testing environment for the development of AI agents for tactical Air Traffic Control (ATC). This includes assessing AI agents using an assessment framework inspired by the Area Basic Training course sat by trainee Air Traffic Control Officers (ATCOs) at NATS's ATCO training college. Additionally, Project Bluebird is aiming to perform the world's first live shadow trial of a sector of airspace, and multi-agent, multi-sector trials on real-world sectors. In~\citet{DTpaper} we argue that these aims necessitate the development of a probabilistic Digital Twin to provide AI agents with an interactive virtual representation of the en route UK airspace and air traffic as well as the artificial airspaces used by NATS for ATCO training. The Digital Twin employs a probabilistic, physics-informed machine learning (PIML) model of aircraft performance to simulate aircraft trajectories in real-world sectors \cite{hodgkin2025probabilisticsimulationaircraftdescent}. The model reflects the inherent uncertainty observed in aircraft performance in a way that can be updated in real-time. There are various components for connecting to agents including an application programming interface (API), a gym for Reinforcement Learning, and a package to measure real-time safety and efficiency metrics. It can generate synthetic training scenarios via prompts to Large Language Models (LLMs), allowing augmentation of training sets for AI agents \cite{gould2025airtrafficgenconfigurableairtraffic}.

The Digital Twin developed by Project Bluebird has been primarily designed as a development platform for AI Air Traffic Control agents. However, there are many potential future use cases for the Digital Twin within ATM. For instance it could be adapted into a training tool for human trainee ATCOs, or form the foundation for operational tools supporting strategic planning, airspace redesign, and conflict detection and resolution. Pursuing any of these use cases would require fostering trust with key stakeholder groups beyond the research team. In training contexts, this would include end users (namely ATCOs), whilst for operational settings it would also include regulatory bodies. It is useful, therefore, to have a framework that can support various forms of stakeholder engagement throughout the life-cycle of a product from research and development to certification and ultimately deployment.

Whilst many components of this Digital Twin could fall under the usual standards of software verification and validation, some of the more novel components do not. These include the planned capability for live data shadowing, LLM-powered synthetic scenario generation, the implementation of machine learned models for probabilistic trajectory prediction (TP), and the Digital Twin's interoperability with AI agents.

As we see it, the challenge is twofold: 
\begin{enumerate}
    \item to offer a level of assurance that can be configured towards a particular application and stakeholder group; and
    \item to flexibly account for AI-enabled systems, such as probabilistic Digital Twins, where the guidance is rapidly evolving and currently incomplete.
\end{enumerate}

To address these related concerns, we have used a methodology and tool known as the Trustworthy and Ethical Assurance (TEA) platform \cite{burr2024}. Central to this methodology is the idea of an assurance case: a structured argument that shows how a set of claims and evidence justify the validity of a top-level Goal Claim.

Assurance cases have been used in safety-critical systems for several decades, and are recommended practice for establishing and assuring safety requirements by the UK's Civil Aviation Authority \cite{CAA_CAP670}. They have, more recently, been extended to account for AI Safety concerns \cite{habli2025bigargumentaisafety}. However, there is no a priori reason why they should only be used for assuring safety and security. As such, the TEA platform seeks to extend this to encompass other areas of assurance, such as explainability, fairness, sustainability, and accuracy. Key to this approach is an open and community-centered focus, whereby researchers, who are not typically involved in the development of assurance cases, can work collaboratively with other stakeholders to identify gaps and co-create best practices for emerging technologies, such as AI-enabled Digital Twins \cite{burr2024, habli2025bigargumentaisafety}. This extension is in line with one of the challenges outlined in EASA's concept paper on guidance for ML applications, namely ``adapting assurance frameworks to cover the specificities of identified AI techniques and address development errors in AI-based systems'' \cite{easa_guidelines_2024}. However, in establishing this community of practice, key capabilities need to be established because researchers do not necessarily have first-hand experience with formal methods of verification and validation used by industry and required by regulators.

By using the TEA platform as an organising structure, we hope to contribute to this growing community of practice by offering a real-world example of an assurance case focused on the goal of assuring sufficient accuracy and fidelity of a Digital Twin that supports the training and evaluation of AI/ML agents for ATC. We highlight the assurance of novel components of the Digital Twin that do not fit easily within traditional frameworks of verification and validation including probabilistic trajectory generation and LLM-driven scenario generation. We connect the assurance case to the relevant emerging literature on AI/ML in ATM and Digital Twins. Given that the current use case for such a system is training and development, rather than operational usage, we aim to assure accuracy and fidelity instead of safety\footnote{However, it should be noted that were such a system to be deployed operationally, safety would be critical but would still depend on an argument over sufficient fidelity and accuracy.}. This will serve as a road map towards demonstrating that a Digital Twin is fit-for-purpose and as a foundation for further intended uses.

\section{Methods}
\label{sec:methods}

\subsection*{Trustworthy and Ethical Assurance}

TEA is a methodology and set of tools for helping teams embed assurance into their existing activities. Key to this is the production of an assurance case: a structured representation (i.e. document) of a reasoned and justified argument to support the validity of a top-level Goal Claim. The TEA platform \cite{tea_platform} is an open-source tool that allows teams and users to collaboratively build such an assurance case using a process of argument-based assurance\footnote{There are various standards for argument-based assurance, such as Goal-Structuring Notation (GSN) and Claims, Argument, and Evidence (CAE), and also meta-frameworks such as the OMG's Structured Assurance Case Metamodel. The TEA platform currently focuses on GSN, but abstracts away from the specific guidance to help users create valid cases without needing to know the detailed guidance \cite{gsn2021}.}. The following core elements define the building blocks of an assurance case:

\begin{itemize}
    \item \textbf{Goal Claim:} a top-level claim about the project or system being developed that helps focus the argument (e.g. `the system is sufficiently safe for its use case').
    \item \textbf{Context:} key descriptions about the context of use to help constrain the scope of the argument (e.g. defining what the system is, who will be using the system, and what the intended environment is).
    \item \textbf{Strategy:} these elements do not make substantive claims themselves, but instead help provide additional structure to the argument by grouping sets of claims and evidence into specific strands (e.g. a sub-argument over a specific component of the over-arching Goal Claim).
    \item \textbf{Property Claim:} lower-level claims about key features of the project or system, which help operationalise and ground the Goal Claim in specific properties (e.g. claims about technical properties of the system, goals about the governance that surrounds the system's design and development).
    \item \textbf{Evidence:} the validity of a claim depends on whether there is sufficient evidence to justify its acceptance. Evidential artifacts will often be produced naturally over the course of a project (e.g. results of verification and validation), and may serve to justify multiple claims.
    \item \textbf{Assumption:} some claims may rest on key assumptions (e.g. that a specific technique is effective for evaluating the performance of a model). Making these assumptions explicit can help internal and external stakeholders assess the overall validity of an argument, or offer additional suggestions to help address gaps.
    \item \textbf{Justification:} similar to Assumptions, justifications help make clear the reasoning of the team producing the assurance case (e.g. why a specific evaluation metric was chosen).
\end{itemize}

\subsection*{Developing the Assurance Case}

We start by articulating the Goal Claim of our assurance case. This is a focused claim (or proposition), to which we then add Context to help define our aims and intended use case. Context is a critical precursor to guiding what Property Claims and Evidence are appropriate for a given Digital Twin. Whilst fidelity and accuracy may have generic requirements across Digital Twins, the choice of what metrics and thresholds to use is often specific to such contexts.

After defining the Goal Claim and Context, we devise a set of strategies and preliminary Property Claims, guided by both the Context and the relevant emerging and established guidance. For Digital Twins, this includes the Gemini Principles along with perspectives on the added challenges of validation generally \cite{thelen_comprehensive_2023, wagg2025philosophical} and for Digital Twins within aviation specifically \cite{ferrari_digital_2024}. We leverage guidance on metrics for trajectory prediction (e.g.\ \citet{mondoloni_assessing_2005}) and validation in safety-critical autonomous systems (e.g. \citet{kochenderfer_algorithms_2025}). We also refer to the well-established Data Quality Dimensions devised by the Data Management Association (DAMA) \cite{DAMA}.

For the components of the assurance case relating to AI or ML, we are especially guided by the AI assurance sections of the EASA Concept Paper on Guidance for Level 1 \& 2 Machine Learning Applications \cite{easa_guidelines_2024} (which we shall henceforth refer to as EASA Guidance). This guidance is both detailed and likely to align with the developing guidance of the CAA, which regulates air traffic control in the UK. It includes draft objectives, means of compliance (MOC) and the scope over which they apply. These often fall into the realms of either documentation and traceability (which we address mostly in Additional Context), and verification and validation (which we address in the Property Claims). The relevant EASA Guidelines are determined by both the level of automation and the end users, and they include sections covering aspects from security to ethics. Given the current focus on accuracy and fidelity, various sections on ethics, security, safety and explainability are excluded from our analysis. These would form the basis of an extended assurance case, which starts to encompass additional goals beyond accuracy and fidelity. Such hierarchically-nested assurance cases are possible via use of modules \cite{habli2025bigargumentaisafety}, but this is beyond the scope of our current work. It should also be noted that the aim here is currently not to meet the future requirements for certification, but to find overlapping objectives based on the Context.

Rather than describe the complete assurance case here, we provide an overview of the top-level Property Claims for each Strategy alongside two `deep dives' into specific Property Claims. These deep dives, presented in shaded boxes within Section \ref{sec:results}, include examples of Evidence and the Assumptions and Justifications that would be sufficient for a given set of Evidence or Property Claim. The reason for this approach is two-fold: first, we aim to focus on the challenging and novel examples that leverage AI/ML whereas a comprehensive Strategy would require an extensive appendix; and, second, the assurance case is still a work in progress. We are creating a living document using the TEA platform and intend to publish the entire assurance case as a Case Study within it.

It is important to note that we do not yet claim that our Digital Twin has been assured for its intended use case. Instead, it serves as a map of the Evidence that must be generated and the thresholds each piece of Evidence must meet.

\section{Results}
\label{sec:results}

Our Goal Claim (G1) is `The Digital Twin of en route UK airspace and Basic Training has sufficient fidelity and accuracy for its intended uses'. This captures the system we are trying to assure, what aspect we are assuring and how it is being applied.

\subsection*{Context}

\begin{figure}
\centering
\includegraphics[width=1\textwidth]{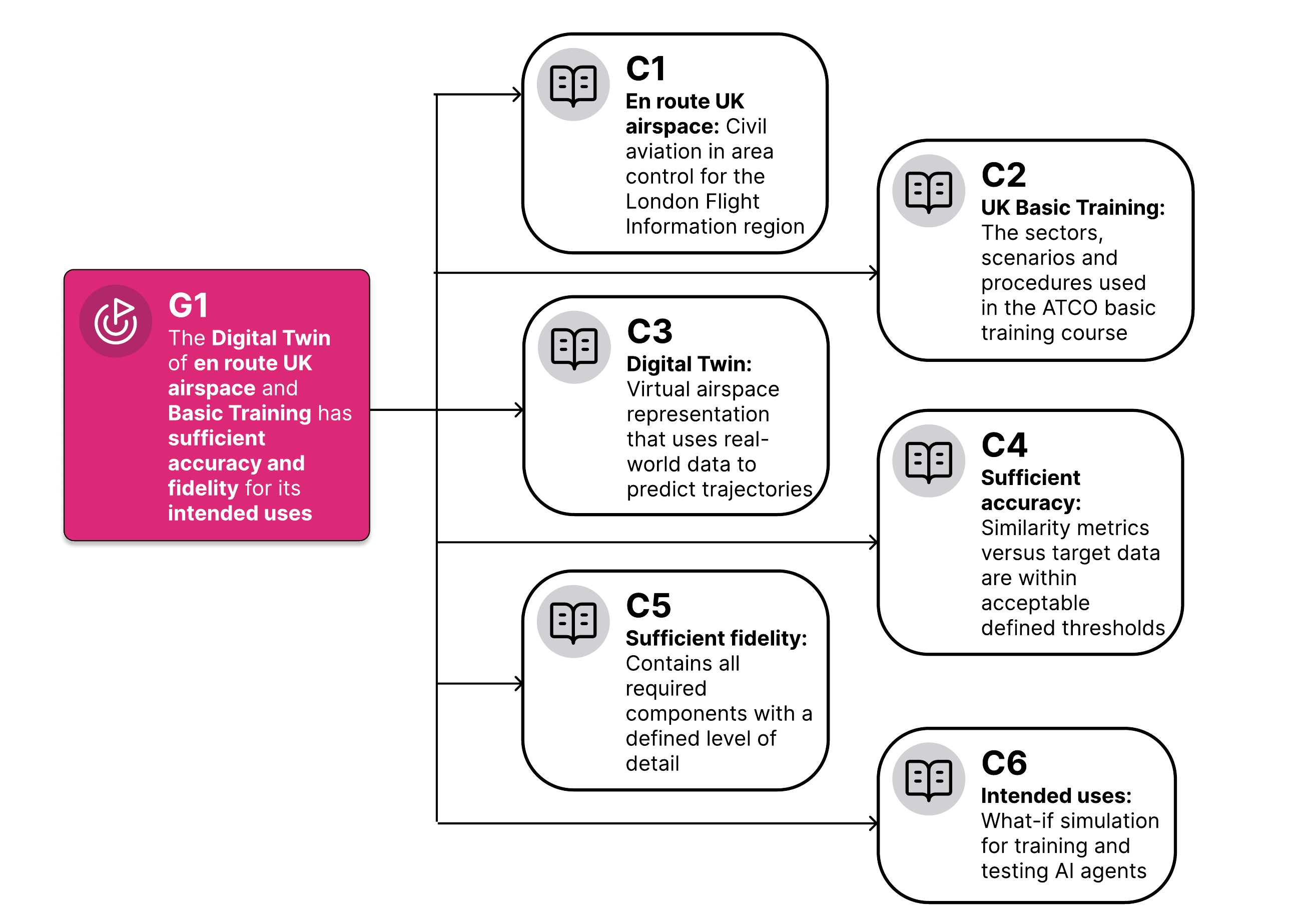}
\caption{Context for the assurance case. G=Goal Claim, C=Context.}
\label{fig:context}
\end{figure}

Figure \ref{fig:context} shows the Context for our Goal Claim, namely defining: `en route UK airspace' (C1), `UK Basic Training' (C2), `Digital Twin' (C3), `sufficient fidelity' (C4), `sufficient accuracy' (C5) and `intended uses' (C6).

\subsubsection*{C1. En route UK airspace}

The Digital Twin models the en route controlled airspace of London Area Control Centre (LACC). This airspace covers altitudes between flight levels 195 and 660 (19,500 ft to 66,000 ft) \cite{CAA_airpace_class}. At present, the more dynamic airspace of London Terminal Control Centre (LTCC), a lower altitude controlled airspace volume that handles departures and arrivals for major London airports \cite{airspace_designation}, is represented in a simplified form within the Digital Twin. Furthermore, we are focused on civil aviation, meaning that aircraft operated by the military and police are out-of-scope, except insofar as they affect civil traffic. This also rules out lower flight levels, aerodromes and uncontrolled airspace regions, implying that, in general, drones and helicopters can be ignored. It also implies that many procedures (e.g.\ SIDs, STARs, speed limits) and clearance types associated with Terminal Control do not need to be accounted for.

\subsubsection*{C2. NATS ATCO Basic Training}

The Digital Twin also models Basic Training environments. This includes the training sector (known as Medway) geometry and an emulator of the trajectory generators used by NATS's ATCO training college, along with the key data which applies to it.  It also includes simplified `I', `X' and `Y' geometries along with a detailed synthetic sector for the purposes of agent training.

\subsubsection*{C3. Digital Twin}

The Digital Twin creates a virtual representation of the LACC environment. It connects to and transforms curated historical radar, clearance, available flight data and weather data. This curated data is the endpoint of a data pipeline which gathers raw historical operational data from sources such as the iFACTS air-traffic management system \cite{nats_ifacts} and post-processed wind field data from the European Centre for Medium-Range Weather Forecasts (ECMWF) \cite{ecmwf}. This data is virtually represented as attributes of software classes that encompass key components of such an environment, including the airspace (e.g.\ sectors and fixes), aircraft (e.g.\ type and position) and events (e.g. clearances and radar blips). `Replay' mode allows the Digital Twin to capture these data as trajectories within its environment. However, in order to efficiently and practically represent this data, there is some level of approximation. For example, Mode S radar surveillance data arrives in discrete `blips' at 6 second intervals. The Digital Twin also performs trajectory prediction (TP): it uses an algorithm that predicts future locations of aircraft by generating a trajectory using the last position of the aircraft and conditioned on contextual information such as the aircraft's filed flight plan and previously issued clearances. Depending on the use case requirement, there are a range of predictors the Digital Twin can deploy of varying speeds and fidelity, including a probabilistic ML model that captures epistemic uncertainties in aircraft performance and the industry state of the art Base of Aircraft Data (BADA) model \cite{nuic2010user}. Capability to handle live one-way data is currently being developed.

The EASA Guidelines \cite{easa_guidelines_2024} define Level 1 AI as systems that provide assistance to humans, Level 2 as teaming with humans and Level 3 as advanced (partially or fully unsupervised) automation. By this definition, the AI agents developed as part of Project Bluebird fall between Levels 2 and 3: they can perform automatic selection and implementation of actions within some spectrum of human override. The Digital Twin itself would arguably meet the definitions of a Level 1A (human augmentation) insofar as it provides information on how proposed clearances affect airspace through trajectory prediction.

\subsubsection*{C4, C5. Sufficient fidelity and sufficient accuracy}

We seek to assure the accuracy and fidelity of the Digital Twin. This is essential for fostering trust that knowledge gleaned from the Digital Twin can be reliably transferred to real-world contexts. For the Gemini Principles, this primarily falls under the scope of `Trust', in particular the subset of `Quality' (must be built on data of appropriate quality).

Accuracy can be defined as how closely the Digital Twin matches the ground-truth data, which would generally involve using appropriate metrics to assess this and a specification to indicate whether or not the metric(s) are within an appropriate threshold. For a distribution of data, we also must decide how to summarise the data, e.g. whether to use the mean or a measure of statistical distance that accounts for higher order statistical moments such as the Kullback-Leibler divergence.

Fidelity can be defined as how realistically the en route UK airspace is represented. This encompasses a number of concepts, such as completeness (are we representing all relevant components and their attributes), resolution (is each component represented to the requisite level of detail), and plausibility (is there a physical law or expected behaviour that is violated). How these are measured is highly dependent on context and can overlap with the same techniques used to calculate accuracy.

More broadly, both fidelity and accuracy rely on the ability to use data or run a model without error in the first place (i.e. having integrity). Given that the Digital Twin is probabilistic, an additional factor to consider is uncertainty quantification. This can impact both fidelity (do the output distributions for a given set of trajectories resemble the real-world, i.e.\ do they have the same diversity) and accuracy (for a given set of predictions, how often do they include the real trajectory).

\subsubsection*{C6. Intended uses}

The Digital Twin has the potential for many different uses: it could be used to plan or optimise flight routes, provide an estimate of complexity and generate new scenarios for training. However, the use case we assure here is its ability to provide what-if estimations for a particular course of actions via its trajectory generator for the purposes of training and testing multiple AI agents operating on virtual representations of real-world sectors and the NATS ATCO college Basic Training environment. Thus, it has an API and gymnasium environment designed to be agent agnostic.

\subsection*{Strategies}

\begin{figure}
\centering
\includegraphics[width=1\textwidth]{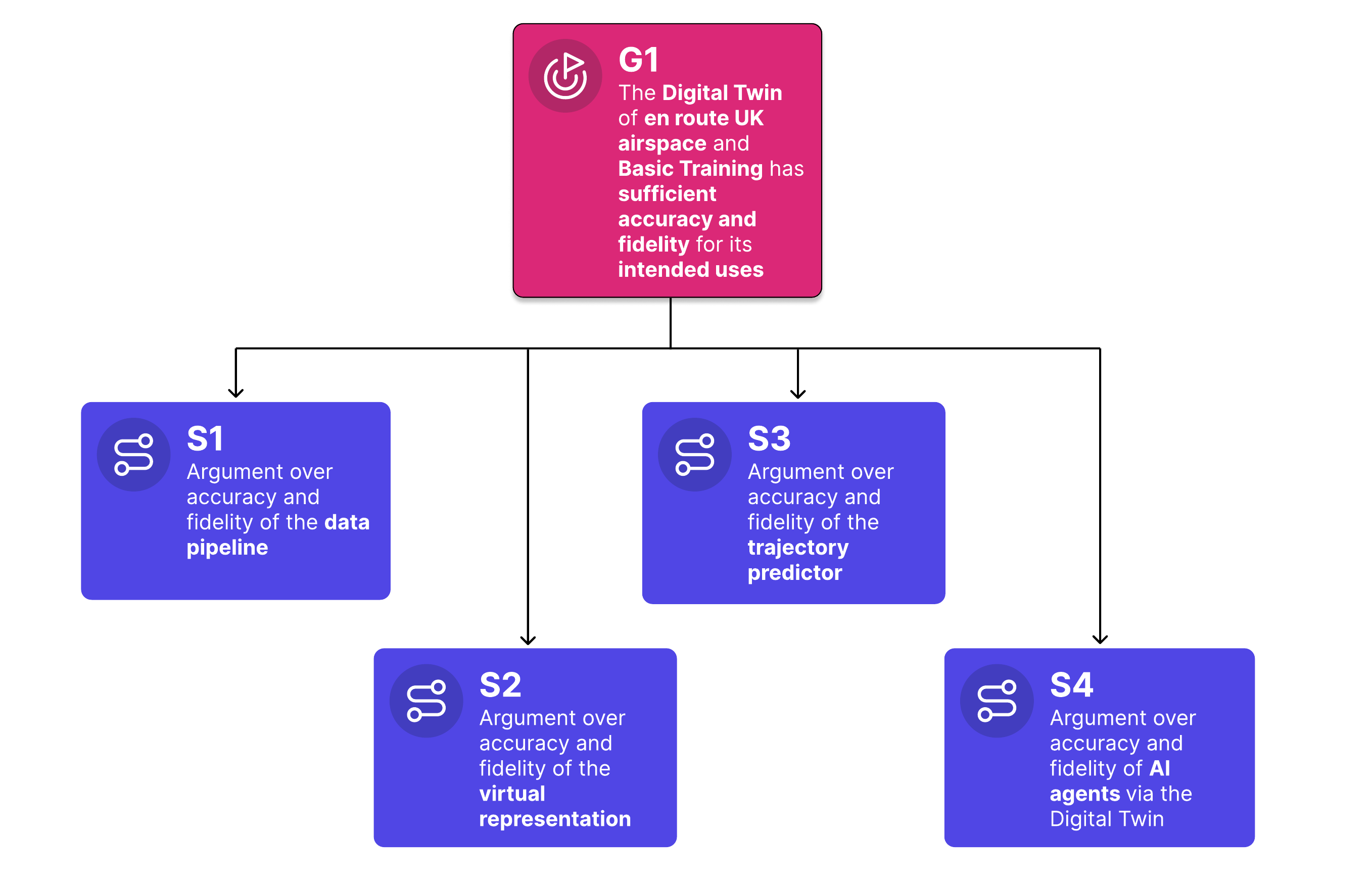}
\caption{Strategies for the assurance case. G=Goal Claim, S=Strategy.}
\label{fig:strategies}
\end{figure}

Figure \ref{fig:strategies} shows the selected Strategies needed to satisfy our Goal Claim. As outlined in the Context, the Digital Twin transforms various raw data into curated datasets via a pipeline, which are in turn converted into a virtual representation, including trajectory replay and prediction. These representations and predictions are used by the AI agents to inform decisions. This suggests separating into Strategies around:
\begin{enumerate}[label=S\arabic*)]
    \item Accuracy and fidelity of the data pipeline and its accuracy versus the real-world.
    \item Accuracy and fidelity of the virtual representation of the data and accuracy of its replay versus the input data.
    \item Accuracy and fidelity of the trajectory predictor and its accuracy versus the replay.
    \item Accuracy and fidelity of AI agents as enabled by the Digital Twin.
\end{enumerate}
This allows us to change our perspective of what our `ground-truth' comparator is, with the caveat that each of these Strategies is conditioned on the previous one.

\subsection*{Strategy 1: Argument over the accuracy and fidelity of the data pipeline} \label{sec:virt_env}

\begin{figure}
\centering
\includegraphics[width=1\textwidth]{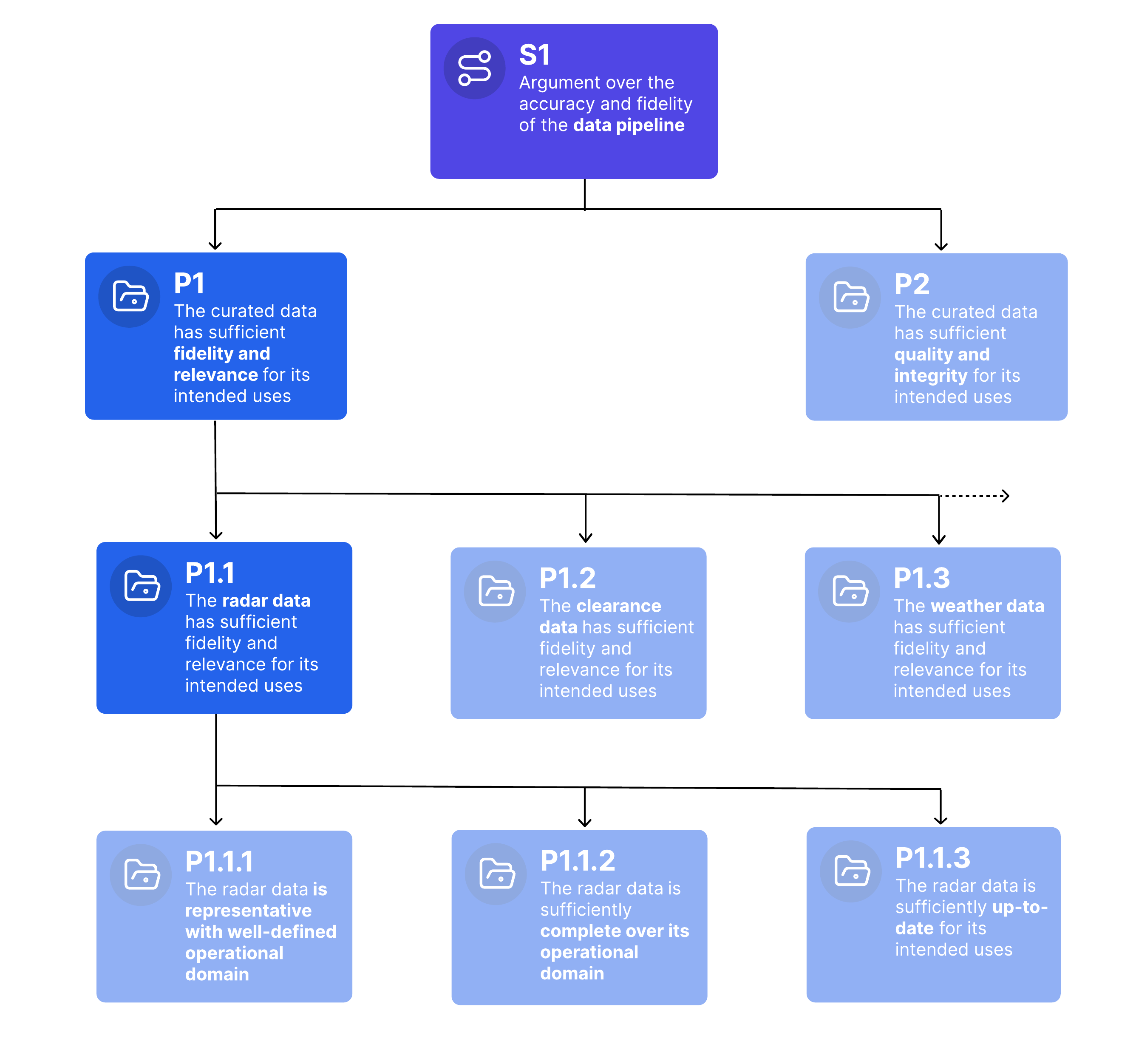}
\caption{Abridged argument over the accuracy and fidelity of the data pipeline. S=Strategy, P=Property Claim.}
\label{fig:data-pipeline}
\end{figure}

Figure \ref{fig:data-pipeline} shows selected elements from the argument over the data pipeline. We provide additional context and then elaborate upon the top-level Property Claims.

\subsubsection*{Additional context}

The Digital Twin requires various sources of historical data to represent UK airspace with sufficient fidelity. Area Control is divided into 3D volumes of jurisdiction (i.e.\ sectors) which are populated with navigation aids and waypoints (i.e.\ fixes) and routes. Information regarding the sector geometry and fix locations is provided by a combination of the Aeronautical Information Publication (AIP) and NATS systems. To obtain aircraft position information for a given flight, such as location, altitude, heading and speed, we use radar track data from ARTAS, a multi-radar tracker system developed by EUROCONTROL \cite{eurocontrol_artas}, which also presents Mode-S data like selected flight levels. Other important information includes flight plans, supplying details on the planned routes, altitudes, speeds and aircraft type. Records of clearances issued by ATCOs to pilots and coordinations (agreements on how and where aircraft pass from one sector to another) are also required. Flight plan data, clearances and sector coordination records come from NATS's proprietary air traffic control systems such as NERC/iFACTS \cite{nats_ifacts}. Meteorological data is provided by forecasts from the UK Meteorological Office (UKMO), and/or ERA-5 reanalysis datasets from the European Centre for Medium-Range Weather Forecasting (ECMWF) \cite{ECMWF_ERA5}. These historical data sources are collected on an ongoing basis from 2016 and encompass over 20 million flights.

The Digital Twin also incorporates data from the Basic Training exercises performed by trainee ATCOs. This includes the geometry of the Medway training sector and its fix locations, as well as formative and summative scenarios, which includes when and where different aircraft appear, any events that occur (e.g.\ radio failure), and wind configurations. Logs from individual student simulations provide data on aircraft trajectories and clearances issued. The aircraft performance parameters for their flight simulator are also included. All of this information comes from the ATCO training college and includes data on all Basic Training exercises from 2024 to 2025.

Recently, the project has gained access to a live-data stream of operational data, enabling real-time shadowing of the live operational situation in the Digital Twin. Live data is sourced from operational ATC systems, in near realtime. The exception is weather data, which will use operationally sourced UK Meteorological Office forecast wind and temperature fields, using either 3-hourly pressure level forecasts or hourly flight level forecasts from the 0.25 degree World Area Forecast System (WAFS) \cite{ukmo_SADIS}.

In order to process the various historical data sources and amalgamate them into formats that can be consumed by the Digital Twin, we have developed a series of data pipelines. The pipelines are hosted in Microsoft Azure to ensure the data is transformed at scale to standard formats. This uses a typical medallion architecture and data engineering services to perform the necessary transformations to the final project formats. The cloud also facilitates the application access control features to the datasets to satisfy NATS security process requirements, which would be very challenging at this scale with typical data transfer methods.

Live data requires separate pipelines and curation due to differences between live and historical data sources and formats. The performance of the operational interfaces imposes additional complexities for data preparation and presentation to the Digital Twin input API and subsequent consumption within the twin itself, whilst maintaining stable low latency and consistent throughput.

\subsubsection*{Property Claim 1: The curated data has sufficient fidelity and relevance for its intended uses}

We need to assure the data the Digital Twin receives is representative and relevant, in line with the EASA Guidance for data management (Objective DA 07-02). This Property Claim should be performed over each dataset, such as radar (P1.1), clearance (P1.2) and weather data (P1.3).

Each dataset should be within the domain in which we are seeking to guarantee accuracy and fidelity (e.g. P1.1.1 for radar), namely UK en route airspace and Basic Training scenarios . This is commonly referred to as the operational domain (OD). A list of requirements for what this should consist of is based on the Context and forms a key piece of evidence. For example, because the Digital Twin is designed to reflect en route airspace (C1) we trim our real-world radar data accordingly to remove other areas (e.g.\ terminal control). We also do not wish to use data with emergency, military or police aircraft. For live shadow mode, real-time monitoring of whether the data stays within the OD is critical. Having predefined bounds allows the Digital Twin to alert agents when its predictions are no longer to be trusted in agreement with EASA Guidance (Objective EXP-05) and best practice \cite{mondoloni_assessing_2005}. This could allow AI agents to rescind control in a future operational environment.  As well as having explicit conditions, we could apply clustering to define the range and density of data we currently have. However, there are tens of millions of instances in our data, and direct use of a model to classify data points as in/out of domain could be prohibitive in terms of memory and storage. Instead, we may define the data as a union of convex hulls of the clusters \cite{kochenderfer_algorithms_2025}. Flight or weather data can also have a large number of features for a given timestamp \cite{easa_roadmap_2024}. An appropriate technique for dimensionality reduction could be applied but would need a sufficiently small reconstruction error.

We also assure the data is sufficiently complete (e.g. P1.1.2 for radar), which again reflects EASA guidelines (Objective DA 07-01), and is also a Data Quality Dimension. This would include specifying what data is required and why. For example, to emulate a given aircraft we need both the clearances issued by the ATCO and the radar data. This is because radar data is needed to initiate a prediction and to assess its accuracy and the clearance data is needed to correctly action trajectory changes in response to these clearances. The EASA Guidance also indicates the need for sufficient resolution in the data. For example, we expect that radar track blips are typically spaced over 6 second intervals; large gaps would have implications for properly simulating and assessing such trajectories. We also need to ensure that there are enough data points within our OD for the model to generalise. One suggested approach (Anticipated MOC DA 07-01) would be to divide the space into hypercubes defined by the relevant features and ensure that the samples used to train models within the Digital Twin sufficiently fill this space. Another is to use density-based clustering where practicable. It should be noted, however, that the incorporation of physics into our ML model via BADA can mitigate some generalisation errors.

We also need to consider timeliness (e.g. P1.1.3 for radar), which is a Data Quality Dimension and a component of the EASA Guidance (Anticipated MOC DA-04). This relates to whether the time period of the data is applicable for the model use. Although Project Bluebird has access to contemporary data, the Digital Twin will require further development and verification work based on prompt operational data streams over cached or historical data. For the live shadow mode, the data would need to be consistently monitored for freshness, with a lag above some threshold triggering a warning.

These objectives are well aligned with general good data practices, which demand an understanding of dataset provenance, including its origins and lineage, in addition to its subsequent processing and transformation. It is reasonable to presume that the majority of Digital Twin input data will originate from ATM systems. As these systems have a known degree of technical and/or process assurance, this offers grounds for a bounded confidence of input data integrity and hence technical data quality.  However, they do not, of themselves, form evidence of direct data quality from other factors (e.g. input completeness), or of direct utility or usability by the Digital Twin.  Many ATM datasets are not natively compatible with machine readability, or AI training data expectations, and require further treatment.

\subsubsection*{Property Claim 2: The curated data has sufficient quality and integrity for its intended uses}

As elucidated earlier, integrity is a precondition for assuring fidelity and accuracy. For data, it must be in a format that can be used as input to the Digital Twin. EASA Guidelines emphasise the need for correctly formatted data that has not been corrupted while stored, processed or collected (Anticipated MOC DA-04). In practice, another means of compliance, namely the need to define the nominal data, is also critical (Anticipated MOC DA-03). These conditions align with the Data Quality Dimension of Validity, i.e.\ that the data conforms to the range and format expected.

Data quality maps to the accuracy of the Digital Twin. EASA Guidelines require that various sources of error and bias are accounted for such as sensor errors, data cleaning errors and annotation errors (Anticipated MOC DM-07-03). An example of a data cleaning error would be if valid indicated airspeed data was mistakenly removed from the dataset. This indicates the need for a defined and correct approach to handling outliers and novelties (Anticipated MOC DA-03). One such case would be distinguishing two tracks that have been given the same callsign mistakenly. This could result in apparently implausible speeds which might erroneously be filtered out when in fact the data needs to be wrangled correctly to obtain the correct speeds. An example of a common annotation error is a `descend when ready' clearance being mislabeled as `descend now' as a default system behaviour. This could result in a simulated aircraft descending immediately instead of after a delay. Whilst one could ideally check these by pulling out the actual transmission data this would be laborious and infeasible at full operational scales and necessary data depth. The solution we take is to use the available radar data to measure delay time before pilots take action and reclassify accordingly. If they cannot be mitigated, errors should be recorded and propagated. For live shadow mode, a critical exercise is to compare its live-streamed values against historical data for a given time period. If we assume the historical data has been assessed for quality, the live data should then match it within some acceptable threshold.

As well as the Data Quality Dimension of Accuracy itself, we need to consider Uniqueness and Consistency: the data wrangling process should handle and avoid duplication (e.g.\ that the identifier we use to label every flight is not shared by more than one flight) and handle or explain values that appear contradictory (e.g.\ if the calculated and measured flight levels differ significantly). Live shadow mode again presents additional challenges: whereas historical flight data is a static `ground truth', live flight data is dynamic and requires online validation.

More widely, there are existing ATM system functional and performance requirements and assessment frameworks. These could be reapplied to equivalent Digital Twin functions, for example, radar track updates and accuracy performance could be assessed in line with the ATC Surveillance tracker using EUROCONTROL's Surveillance Analysis Support System for ATC Centres \cite{SASSC}. This also highlights two important considerations: firstly that there is potential for re-use of existing requirements and secondly that agents may significantly impact current upstream functional safety and/or performance requirements for current and future ATC systems and overarching aviation system of systems.

The EASA Guidance also allude to the the potential need for software development assurance for data pipelines themselves and not just data verification. Such assurance would need to provide coverage between the originating ATM system or data archive service (including recovery procedures), and the final output data of the pipeline.  There are good-practice frameworks for data management which could be used for this purpose \cite{DAMA, FAIR}.

\subsection*{Strategy 2: Argument over the accuracy and fidelity of the virtual environment}

\begin{figure}
\centering
\includegraphics[width=1\textwidth]{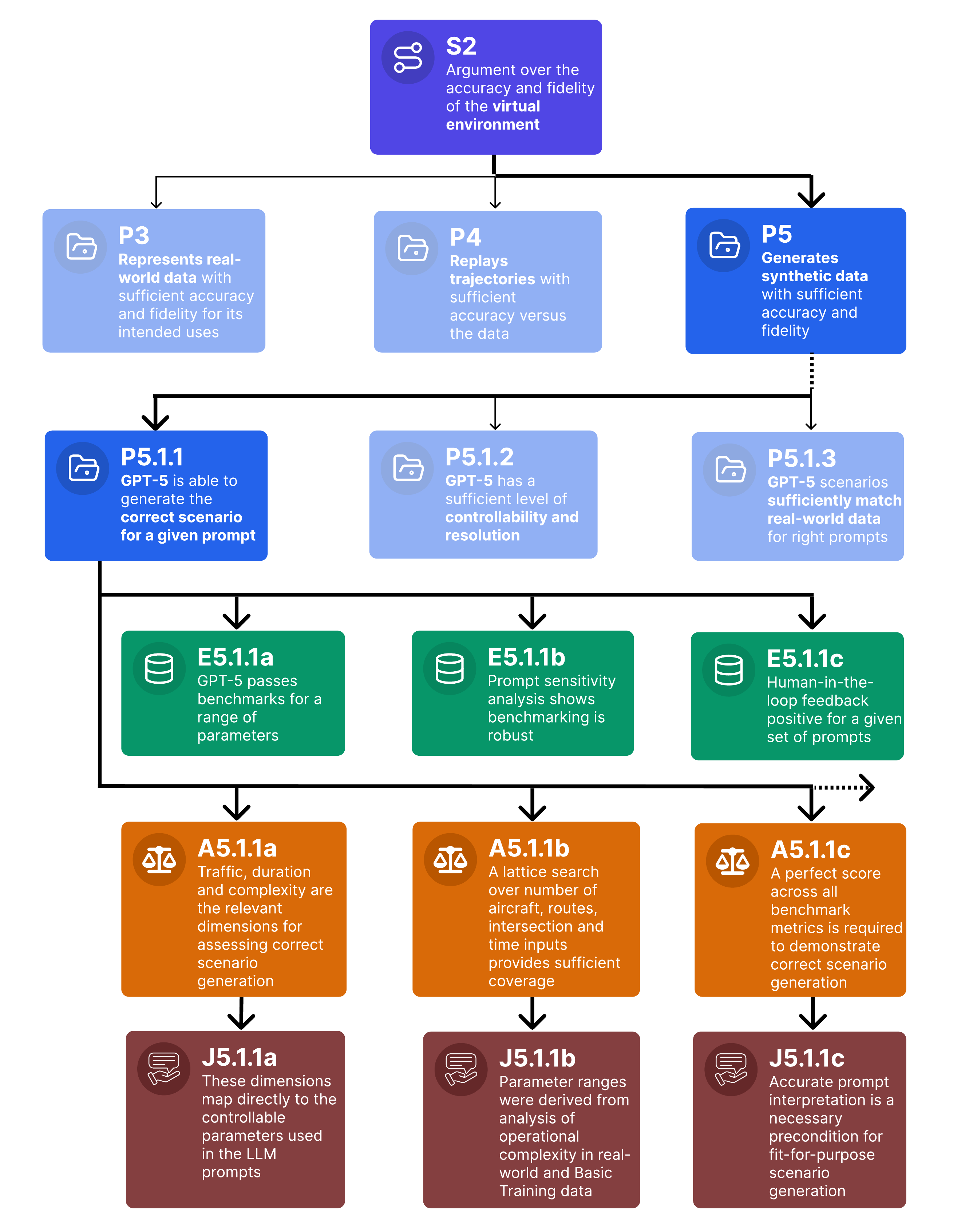}
\caption{Abridged argument over accuracy and fidelity of the virtual environment. S=Strategy, P=Property Claim, E=Evidence, A=Assumption, J=Justification. Deep dive is highlighted by the bold lines and arrows.}
\label{fig:virtual-env}
\end{figure}

Figure \ref{fig:virtual-env} shows selected elements from the argument over the virtual environment. We elaborate upon the top-level Property Claims, and focus on one set of Evidence about benchmarking the scenario generator. For this argument, additional context is provided within each Property Claim, rather than as a standalone section.

\subsubsection*{Property Claim 3: Represents real-world data with sufficient accuracy and fidelity for its intended uses}

The Digital Twin creates a virtual representation of London Area Control or Basic Training populated by operational data from the pipeline. The software used to perform this is an object-oriented approach, currently implemented in Python, with classes representing key components of such an environment, including the airspace, aircraft, events and weather. The attributes of these classes are then instantiated with the data from the pipeline. Each aircraft is itself comprised of attributes that describe the current state of the aircraft (e.g.\ altitudes, locations and airspeeds) and also contain data pertinent to its future intentions such as destination and flight plan information. The environment and each of its constituents are under configuration management (Anticipated MOC CM-01) through version control, dependency management, documented releases and version tracking.

Because there are many components within the virtual representation and many sources of data to instantiate these components, this Property Claim has many sub-claims around airspace, aircraft, weather and events. These, in turn, will have their own sub-claims such as the kind of aircraft types we are representing.

As there are currently no AI/ML components related to any of these representations for real-world data, the V cycle of software development forms a large part of the approach as recommended in the EASA Guidance. This includes specifying requirements for each component, what components are important, and applying multiple unit tests to ensure each one functions as intended. Nevertheless, some of the same data management considerations outlined in the EASA Guidance carry over from the argument over the data pipeline. That includes resolution and how it differs from the curated data, e.g.\ the precision of the locations of fixes, the boundaries of sectors, and the spatial and temporal resolution of the wind field. It also includes documentation of how the data maps from curated data into the model representation, and ensuring these values are consistently and correctly mapped. This includes changes in nomenclature, type conversions or rounding, and any further wrangling of the data. For example, in order to calculate wind fields at any given location we apply a nearest-neighbor imputation of the curated data.

\subsubsection*{Property Claim 4: Replays trajectories with sufficient accuracy versus real-world data}

The Digital Twin has the ability to replay operational scenarios exactly as they occurred using historic radar and clearance data. This replayed data is used as input to the TP, which means that for a given trajectory we can use the Digital Twin in replay mode to evaluate how the reduction in fidelity needed for the virtual representation affects the accuracy versus the curated data. This allows errors arising from inaccuracy in the TP to be isolated from errors caused by the input data and the approximations inherent in the virtual environment (e.g.\ the loss of resolution resulting from transforming Mode S radar surveillance data and clearances from continuous time into discrete 6 second `blips').

The precondition around integrity here is that the replay produces trajectories for each aircraft for which accuracy can then be measured. This includes the avoidance of runtime error, which can be evidenced by measuring the coverage over the data for which the model runs successfully. Integration tests have been developed to guarantee this is the case over particular datasets. Even when the virtual environment runs, we need to ensure each aircraft in the data should be instantiated in the virtual environment, i.e.\ there are no missing aircraft.

For a given aircraft, there are a range of common metrics to measure the accuracy of the inputs to the TP \cite{mondoloni_assessing_2005}. These include:
\begin{itemize}
    \item \textbf{Initial condition error:} measuring the disparity between the coordinates at which an aircraft appears in the replay versus the curated data.
    \item \textbf{Wind vector error:} This is the error arising from approximations in wind data interpolation and quality of the data itself such that the magnitude and/or direction of the wind vector do not match reality. For replays, ground speed is replicated from the data but calibrated airspeed (CAS) is calculated by applying wind vectors to obtain true airspeed (TAS) and then converting TAS to CAS using conversion formulae that assume a standard atmosphere. Therefore, we can directly monitor the error arising from wind via the difference between the the CAS of the replay versus the indicated airspeed (IAS) of the curated data. In our case, other approximations also apply: we assume that instrument error in measurements of IAS are negligible, and standard atmospheric model parameters apply. Thus this is really measuring an IAS$\rightarrow$ground speed conversion error. Similarly, a comparison between the heading$\rightarrow$ground track angle in the Digital Twin needs to be compared to the real-world values.
    \item \textbf{Aircraft attribute mismatch:} This measures any differences in the aircraft attributes in the virtual representation versus the curated data. If the filed route flown by the aircraft was incorrect, for example, this would result in the wrong lateral path being flown in the TP.
    \item \textbf{Action intent error:} The Digital Twin can emulate the most common speed, lateral and vertical clearances issued in London Area Control (see \citet{DTpaper} for a complete list). In each case, we need to assess that the action is correctly implemented (e.g.\ that the selected flight level of an aircraft matches its last vertical instruction). However, holding patterns and conditional clearances (e.g. `Speed 220 knots 12 miles before fix') are not yet supported by the Digital Twin. Furthermore, speed instructions which specify `speed greater than' or `speed less than' are interpreted as `speed equals'. It is important to identify where errors are resulting from this loss of fidelity in how clearances are represented as it rules out the data pipeline or the trajectory predictor as the source of inaccuracy.
\end{itemize}

Some of these errors are irreducible, an example being wind vector error. In such cases, uncertainties should be propagated into our assessments of uncertainty and confidence in TP. This could potentially be via method such as interval arithmetic or more sophisticated methods. Although non-probabilistic methods for modelling epistemic uncertainty are less common in Digital Twin applications \cite{thelen_comprehensive_2023}, they could be complementary if the sources of the uncertainties captured by the probabilistic TP and propagated via non-probabilistic means are independent. Alternatively, a probabilistic prediction of the IAS$\rightarrow$ground speed conversation could be trained. Live data also presents additional challenges: forecast data would need to be used rather than historical weather data and so the discrepancies between these two and how they affect wind error needs to be assessed.

The discretisation of the radar and clearance data into 6 second blips does not affect the accuracy of the TP when compared to the replay (since both use that approximation) but is a source of error compared to the curated data. In order to measure this error, we apply a linear interpolation of the replay radar to each data point of the iFACTS radar data. The resulting distance in interpolated latitude, longitude and flight level is used to quantify the distribution of these discretisation errors.

\subsubsection*{Property Claim 5: Generates synthetic data with sufficient accuracy and fidelity}

The Digital Twin can also generate its own synthetic datasets in lieu of real-world data. This allows generation of additional training scenarios for AI agents, which can augment situations that are rare in the data or test their reactions to unexpected scenarios. These scenarios are generated by transforming our sector into a simplified graphical representation with a discrete location grid. A prompt is then given to a Large Language Model (LLM) to generate a combination of aircraft which conform to user-specified parameters on the number of available routes, scenario duration, number of aircraft and number of interactions (i.e.\ situations in which action must be taken to de-conflict aircraft). Future work plans to incorporate the dynamic introduction of unexpected events such as storms.

Scenario generation needs to be of sufficient accuracy and fidelity for training of AI agents for each LLM. For example, Gemini-2.5-Pro would have a separate Property Claim to GPT-5 (these are omitted from Figure \ref{fig:virtual-env} for brevity but follow the pattern P5.1, P5.2, etc). For a given LLM, the prompts should reliably create scenarios which satisfy the user-supplied parameters (e.g. P5.1.1 for GPT-5), have a sufficient degree of controllability to be useful for the agents (e.g. P5.1.2 for GPT-5), and be able to generate real-world scenarios with sufficient similarity to real-world data (e.g. P5.1.3 for GPT-5).

To determine that it satisfies user-supplied parameters (e.g. P5.1.1 for GPT-5), a given LLM should be able to pass key benchmarks that compare prompt inputs on factors such as number of interacting aircraft with the expected outputs. Tests of robustness to prompts such as sensitivity analysis can be applied. For more complex prompts that are more difficult to automatically verify, it is critical to employ a human-in-the-loop approach (see `deep dive' below).

\begin{tcolorbox}[colback=white, colframe=gray!50!black, breakable]

\textbf{Deep dive on Property Claim 5.1.1\\
GPT-5 is able to generate the correct scenario for a given prompt}\\

For a given prompt, we need to ensure the LLM is able to reliably fulfil a number of performance benchmarks for the specific task of scenario generation (E5.1.1a). A key challenge is to mitigate or identify hallucinations. For non-interacting aircraft, we assume that these benchmarks measure whether the traffic volume, the scenario duration, and the number of route interactions are matched to specifications, and complexity is handled (i.e.\ the LLMs can handle sectors containing routes which intersect more with one another whilst avoiding aircraft interactions) (A5.1.1a). A similar suite of benchmarks are calculated for interacting aircraft. The LLM can accept prompts which specify traffic volume, number of routes and their intersections, and scenario duration. Therefore, the justification for these choices of benchmarks is that they each map to one of the prompts that the LLM can be given (J5.1.1a). To perform these benchmarks, each are varied across a set of parameters, averaged over 10 synthetic sectors. For traffic volume, number of aircraft is varied from 2 to 30 with a fixed value of duration (12 time steps), routes (7) and intersections (7). For other parameters, a similar approach is taken where the key parameter is varied across a range of values whilst the other parameters are fixed (A5.1.1b). The specific values for number of aircraft, number of routes and number of intersections were chosen to replicate the type of operational complexity an ATCO would have to manage (J5.1.1b). Relevant aircraft and airspace configurations in real-world and Basic Training data were analysed to assess what values one could reasonably expect to see. The averaging over multiple sectors ensures statistical confidence in the results.

There is a risk that some scenarios may be rejected as incorrect because they are theoretically unreachable rather than through some flaw in the performance of the LLM. We therefore also verify that all parameters in our benchmarks are solvable. With a feedback loop, GPT-5 is able to fulfill this threshold across all metrics, and is close to meeting this threshold even without the feedback loop. Figure \ref{fig:gpt5-pass} shows an example of GPT-5 passing one benchmark but see  \citet{gould2025airtrafficgenconfigurableairtraffic} for further evidence and details on this work.

\begin{center}
\includegraphics[
    trim=0 170 0 0,
    clip=true,
    width=0.8\textwidth
]{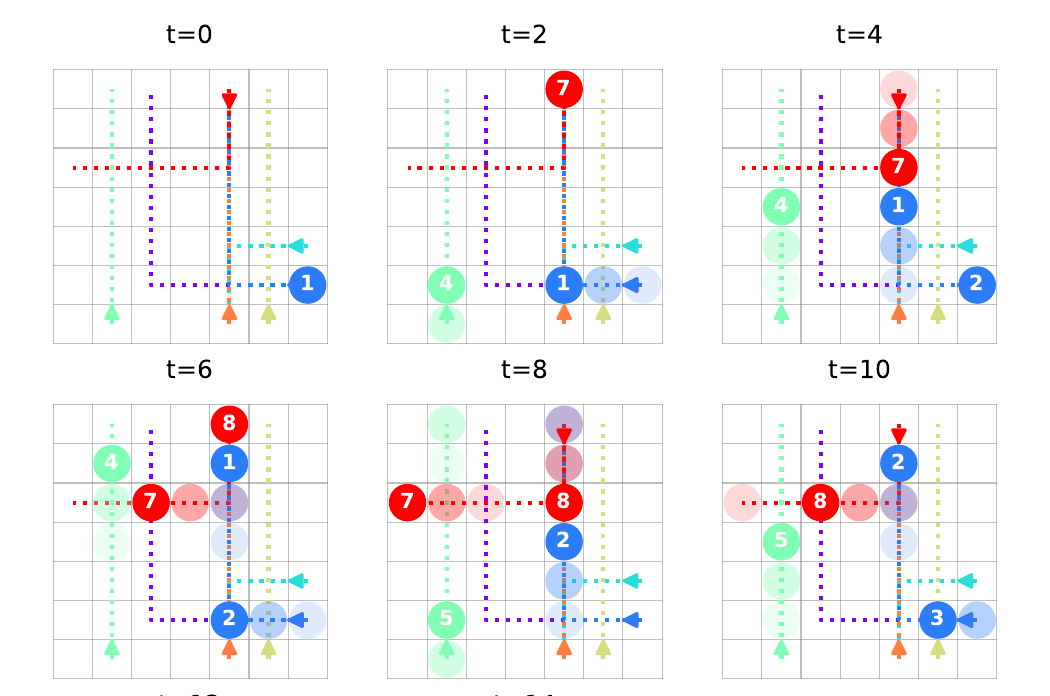}
\captionof{figure}{GPT-5 solving a benchmark with 8 aircraft successfully: there are no interactions in this scenario. Aircraft are never adjacent, which indicates an interaction. This is the simplified grid representation in which each grid point is 20 NMI. Snapshots show two time units from the start of a scenario. Aircraft are colour coded with their current location shown by the solid circle, their trace by the transparent circles, and their paths by the dashed lines.}
\label{fig:gpt5-pass}
\end{center}

\vspace{1em} 

Hybrid AI methods like LLM-driven scenario generation are currently outside of scope of EASA Guidance, although the use of such methods within a regulated training environment would require assurance. However, we use pretrained models with no fine-tuning and implicitly assume that the learning process for these commonly utilised models can be accepted as sufficiently documented and verified elsewhere. Instead, the `learning' process relates to a feedback loop in which the LLMs are given further prompts pointing out where they made an error based on these performance benchmarks. This process can be automated into a conversation which allows them to self-correct and, for some LLMs, to achieve a perfect score across all non-interacting benchmarks after this process. We therefore assume a threshold of 100\% for these benchmarks (A5.1.1c) because accurate interpretation of prompts is a necessary and possible precondition for fit-for-purpose scenario generation without hallucination (J5.1.1c).

The above performance benchmarks apply to a carefully curated set of prompt templates, whose development is documented in notebooks. However, as the prompts are extended to more detailed requests such as the kinds of interactions we want or the aircraft performance, it becomes increasingly important that the LLM can handle permutations in its input. To evidence the robustness of an LLM to a prompt for a given scenario, sensitivity analysis should be performed by seeing how variations in prompt format that maintain that same content do not significantly affect the benchmarking results (E5.1.1b). There are multiple frameworks, such as FORMATSPREAD \cite{schlar_psa_2024} and ProSa \cite{prosa}, which are built for such analysis. However, for more domain-specific variations we should develop a suite of input prompt variations which capture typical nomenclature used in radio transmission. Likewise these more complex prompts (e.g.\ `Create a scenario with two aircraft in trail and a third crossing tracks with these two' shown in Figure \ref{fig:complex-prompt}) are difficult to automatically verify. Instead, we would use a human-in-the-loop approach where ATCOs can visually assess how well a prompt actually succeeded and score key aspects on a Likert scale (E5.1.1c). The feedback and human-in-the-loop constraints on the LLM outputs such that they are truthful conforms to suggested best-practice for the use of LLMs in research as `zero-shot translators' \cite{truthful_llms_2023}.

\begin{center}
\includegraphics[
    trim=0 180 0 0,
    clip=true,
    width=0.8\textwidth
]{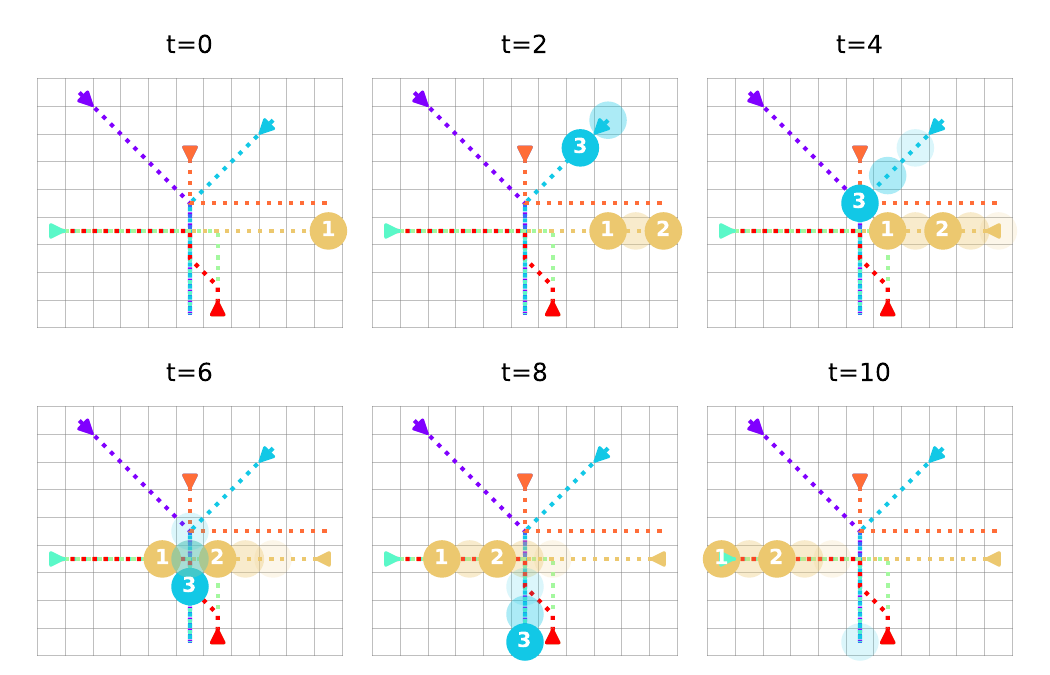}
\captionof{figure}{Example of a prompt to GPT-5 to `Create a scenario with two aircraft in trail and a third crossing tracks with these two'.}
\label{fig:complex-prompt}
\end{center}

\end{tcolorbox}

To determine requisite controllability (e.g. P5.1.2 for GPT-5), similar evidence is gathered by discussions with ATCOs on what constraints need to be imposed and what additional inputs should be allowed for more flexible prompting and scenario generation. This would help to develop a set of specifications which the Scenario Generator would need to check. This also informs the resolution of the simplified airspace representation that the LLM directly uses (which is then transformed into the detailed airspace). For example, the choice of 20 nautical mile grid points for considering whether aircraft are interacting is a trade-off between the experience of ATCOs and what allows the performance benchmarks to be reasonably achieved.

To determine verisimilitude to real-world data (e.g. P5.1.3 for GPT-5), one needs to be able to calibrate the set of prompted inputs which resemble real-world data and then validate against held-out real-world data (e.g.\ different days or sectors). It is not enough that the LLM faithfully fulfils prompts: we need to devise the right prompts to give it. In this case we wish to compare key distributions in the real-world data such as the coordinations, number of routes and airspace complexity for a given set of prompt parameters. The kinds of statistical measures we could use to achieve this are detailed in the argument over the trajectory predictor. Some features could be directly mined from the data pipeline whilst others might require techniques, such as clustering, to determine if the pattern of traffic flows are similar. This can be measured with complexity metrics such as NATS's Traffic Load Prediction Device \cite{complexity_tlpd} or using a graphical neural network approach \cite{complexity_henderson}. Feedback from ATCOs via similar methodologies would again be critical for external validation.

\subsection*{Strategy 3: Argument over the accuracy and fidelity of the trajectory predictor}

\begin{figure}
\centering
\includegraphics[width=1\textwidth]{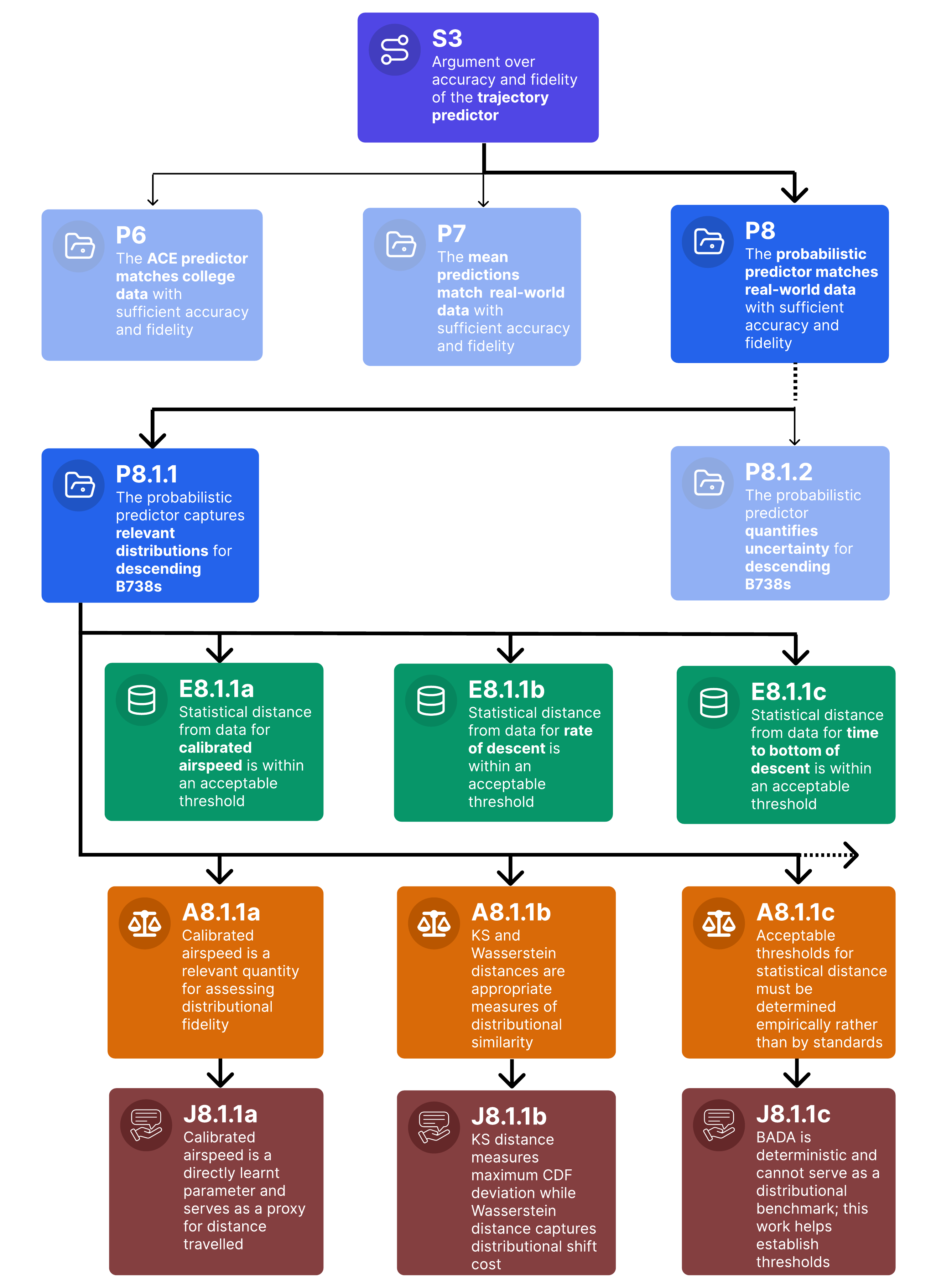}
\caption{Abridged argument over accuracy and fidelity of the trajectory predictor. S=Strategy, P=Property Claim, E=Evidence, A=Assumption, J=Justification. Deep dive is highlighted by the bold lines and arrows.}
\label{fig:traj-pred}
\end{figure}

Figure \ref{fig:traj-pred} shows selected elements from the argument over the trajectory predictor. We provide additional context and then elaborate upon the top-level Property Claims, and focus on one set of Evidence concerning how well the probabilistic predictor quantifies uncertainty.

\subsubsection*{Additional context}

In order for AI agents to make decisions and be assessed, the Digital Twin must be able to emulate real-world trajectories. There are a range of predictors it can use, including the commonly used Base of Aircraft Data (BADA) model. Within BADA, aircraft performance is formulated as a partial differential equation (PDE) that is solved numerically. The PDE is an energy balance equation that relates the rate of climb/descent (ROCD) and true airspeed (TAS) of an aircraft to the relative change in kinetic and gravitational potential energy. In BADA, the ROCD is represented using \cite{nuic2010user}:

\begin{equation}
    \frac{dh}{dt} = \frac{T-\Delta T}{T} \Big[ \frac{(T_{HR}-D)V_{TAS}}{mg_0} \Big] f(M), \label{eq:bada_rocd}
\end{equation}
where $\frac{dh}{dt}$ denotes the ROCD, $V_{TAS}$ the TAS, $m$ the aircraft mass, $g_0$ the acceleration due to gravity, $T_{HR}$ the aircraft thrust, $D$ the aircraft drag, and $f(\cdot)$ the energy share factor, defined as a function of the Mach number, $M$. The energy share factor governs the tradeoff between changes in gravitational potential energy and kinetic energy. It has a different functional form depending on whether the aircraft is flying at constant calibrated airspeed (CAS) or Mach speed. Additionally, $T$ represents air temperature in the International Standard Atmosphere (ISA) model and $\Delta T$ a temperature correction that may be applied. In BADA, parameters are specific to an aircraft type.

The BADA model is deterministic. However, there are a number of epistemic uncertainties in ATM that will introduce variation between predicted and observed trajectories. For example, aircraft mass, pilot response times and performance settings are typically unavailable to the ATCO. It is therefore necessary to generate simulated trajectories probabilistically, such that real-world uncertainty can be captured in simulations using the Digital Twin. We are developing a probabilistic predictor that must perform two functions: firstly, to provide a deterministic trajectory prediction that can be used by either agent or human controllers as a decision support tool; and secondly, to generate probabilistic trajectories that reflect real-world levels of uncertainty.

One component of the probabilistic predictor relates to climbs and descents. Purely data-driven approaches may result in unrealistic trajectories, e.g. aircraft in climb or descent must always have a rate of climb/descent (ROCD) above the minimum rate specified in UK airspace and accelerations, both longitudinal and vertical, should remain within sensible bounds that reflect the operational constraints of passenger aircraft. Therefore, the probabilistic predictor applies physics-informed machine learning (PIML): it uses BADA but with a machine learned correction term applied to the aircraft drag and CAS, denoted $V_{CAS}$, in descent and thrust and CAS in climb ($V_{CAS}$ and $V_{TAS}$ are linked by conversion formulae in BADA). The machine learned corrections (for descent) are functions of the altitude above sea level, $h$, and are of the form:

\begin{subequations} \label{eq:fpca}
\begin{equation}
    {D}(h) = \mu_D(h) + \sum_{i=1}^{n_\alpha} {\alpha}_{i} \phi_i (h) , \label{eq:fpca_drag}
\end{equation}
\begin{equation}
    {V}_{CAS}(h) = \mu_V(h) + \sum_{j=1}^{n_\beta}{\beta}_{j} \psi_j (h), \label{eq:fpca_cas}
\end{equation}
\end{subequations}
where $\mu_D$ and $\mu_V$ are mean functions and $\boldsymbol{\alpha}\in\Re^{n_\alpha}$ and $\boldsymbol{\beta}\in\Re^{n_\beta}$ a set of (uncertain) model weights. The functions $\phi(\cdot)$ and $\psi(\cdot)$ are orthonormal basis functions, satisfying the conditions: 
\begin{equation}
    \int \phi_i(h) \phi_j(h) dh = \delta_{ij}\; \text{and} \;\int \psi_k(h) \psi_l(h) dh = \delta_{kl},
\end{equation}
with $i,j\in[1,n_\alpha]$ and $k,l\in[1,n_\beta]$. These are derived through functional Principal Component Analysis, and the basis functions are discrete and defined over a discretised grid of flight levels. There are equivalent expressions for thrust and CAS in the climb phase. 

A probabilistic model is trained that is specific to the distribution of model weights, $p(\boldsymbol{\alpha}, \boldsymbol{\beta})$, for each aircraft type. \citet{hodgkin2025probabilisticsimulationaircraftdescent} evaluated a number of methods to approximate this distribution, including Gaussian mixture models and normalising flows. In the implementation of the Digital Twin discussed here $p(\boldsymbol{\alpha}, \boldsymbol{\beta})$ is modelled as a multivariate Gaussian distribution for climbs and descents, fitted to training data using expectation maximisation. The interested reader is referred to \citet{Pepper_DataBADA} and  \citet{hodgkin2025probabilisticsimulationaircraftdescent} for more details. Probabilistic cruise speeds are generated by replacing BADA-defined CAS and Mach values with samples from the probability mass function of the training data. Table~\ref{table:predictor_summary} summarises the various predictors that will be discussed in this section.

\begin{center}
\captionof{table}{Summary of trajectory predictors.}
\begin{tabular}{l|p{30em}}
    \hline
    Trajectory Predictor & Description  \\
    \hline 
    Probabilistic & A probabilistic, PIML model of aircraft thrust/drag and airspeed which can be used to generate simulated trajectories that capture real-world levels of uncertainty \\
    Probabilistic (mean-mode) & A predictor that uses the mean thrust/drag and airspeed corrections. Intended as a decision support tool for humans and agents that can be calibrated to aircraft performance within a specific airspace  \\
    BADA & The industry standard TP model, which we use to benchmark the capabilities of the Digital Twin \\
    ACE & The in-house trajectory predictor used at NATS's ATCO training college, which is replicated within the Digital Twin \\
    \hline 
\end{tabular}
\label{table:predictor_summary}
\end{center}

\subsubsection*{Property Claim 6: The ACE predictor matches college data with sufficient accuracy and fidelity}
The simulators used at NATS's ATCO training college employ an in-house predictor, called ACE (Table \ref{table:predictor_summary}). Because the Digital Twin aims to emulate Basic Training, it is important to ensure this predictor has sufficient accuracy and fidelity. Furthermore, because we have access to NATS's proprietary codebase for this predictor in addition to data from the pipeline, we can use this as a de facto verification of the generic predictor components in the Digital Twin. That is, because we have a set of known solutions for a given set of simulations we can ensure there are no unacceptable discrepancies resulting from numerical approximations and transpiling the code.

For a given aircraft, there are a range of common metrics to measure the accuracy of the trajectory emulations \cite{mondoloni_assessing_2005}. We focus on one that captures the mean absolute error (MAE) each spatial dimension:
\begin{itemize}
    \item \textbf{Along-track error}: measures the difference in the predicted horizontal location of the aircraft and the actual location at a given time, projected onto the direction parallel to the heading of the replayed aircraft. One contributor to this error could be an incorrect ground speed.
    \item \textbf{Cross-track error}: measures the difference in the predicted horizontal location of the aircraft and the replayed location at a given time, projected onto the direction perpendicular to the heading of the replayed aircraft. One contributor to this error could be an incorrect filed flight plan.
    \item \textbf{Vertical error}: Measures the difference in the predicted flight level of an aircraft and the replayed flight level at a given time. One contributor to this error could be that the aircraft actions a "level by" clearance at the wrong time.
\end{itemize}
It is useful to consider an overall measure of distance across all three dimensions, but it is important to note the different scales they operate at in ATM: in UK en route airspace, aircraft separation must be either 5 nautical miles horizontally or 10 flight levels laterally. Principal component analysis of en route radar data could help give an approximate weighting to any average across the three dimensional components.

Because the virtual environment discretises time into uniform blips, we can directly calculate each of these distances for each given time point rather than needing to account for errors in time as well. This has already been accounted for in the Argument over virtual environment (S2). We then find the maximum of these along the length of each trajectory to highlight aircraft which are incorrectly simulated. For our purposes, these are aircraft that exceed some threshold value across multiple simulations for a given scenario. Fidelity is ensured in a similar manner to with the virtual environment: we develop a set of specifications for what features of the Basic Training simulator we should emulate and at what level of resolution. For example, we would like to emulate all the clearances which relate to area control but rule most of those relating to terminal control out of scope.

\subsubsection*{Property Claim 7: The mean predictions match real-world data with sufficient accuracy and fidelity}

The probabilistic trajectory predictor may be used in a deterministic mode by sampling the most likely trajectory, i.e. setting $D = \mu_D (h)$ and $V_{CAS} = \mu_V$ and solving \eqref{eq:bada_rocd} numerically to provide a deterministic prediction of the future position of the aircraft that can support decision making.

For such AI/ML components, the EASA Guidance advocates for its W-shaped assurance process, which is a modification of the V-cycle that captures the learning process. This includes documentation of model selection and parameters (Anticipated MOC DA-01), which we give in the Additional Context and \citet{hodgkin2025probabilisticsimulationaircraftdescent}. It also includes independent validation, test and train sets (Objective DM-07-05), which in our case is a standard 80:20 train-test split with a 5-fold cross-validation for explained variance selection. The choice of training data for a given ML/AI component and the conditions under which it operates may be a subset of the OD known as the Operational Design Domain (ODD). The EASA Guidance notes that the same objectives and modes of compliance around OD (i.e.\ in our argument over the data pipeline) apply here \cite{easa_guidelines_2024}. This would include the training period of the data (currently July to September 2019) and further preprocessing (e.g. only including aircraft in a descent above the minimum rate in UK airspace of 500 ft/min). A separate predictor is trained for each aircraft type e.g. Boeing 737-800 (B738). By the same logic, verification of the learning process must assess any parameters relevant to that process. We calculate, for example, the absolute error between the mean calibrated airspeed of the predictor and the data. These errors should be superior to BADA predictions as a minimum threshold.

The W-shaped assurance process also includes inference model verification. We can take a similar approach as Property Claim 1: measure the accuracy of mean-mode predictions against a wider set of real-world data using the MAE of the along-track, cross-track and vertical errors for a given aircraft. However, the college simulations of Property Claim 1 have no epistemic uncertainties, which means a single prediction can be made along an entire prediction. In contrast, real-world predictions diverge over time due to such uncertainties. Consequently, the errors need to be calculated across a time horizon or look-ahead time after which all values are reset to match replay and then another prediction is made. An error point is thus calculated between the prediction and data after each look-ahead time interval. The choice of look-ahead time is an inherent Assumption, and we choose two intervals:
\begin{enumerate}
    \item One radar blip (i.e.\ 6 seconds), which is the minimum interval in the virtual environment. This is a fixed time which unambiguously can be matched to at most one action.
    \item The interval from one clearance to the next, varies in duration. This allows one to diagnose the impact over the natural horizon.
\end{enumerate}
The mean errors for a given look-ahead time are compared to BADA and, as a minimum threshold, should be superior to it according to a level of statistical significance (e.g.\ above some defined Bayes factor). 

It is critical not only to demonstrate that the predictor has sufficient accuracy globally (i.e.\ across the entire ODD) but also at a local level. For example, we need to know if it performs poorly for a given aircraft type. Likewise, there would be separate higher-level claims for descending aircraft, climbing aircraft and so on. Thus we would sub-divide our claim (not pictured) into elements over different maneuvers, clearances and routes, which are further sub-divided by different aircraft types (e.g. `the predictor matches real-world data with sufficient accuracy and fidelity for descending B738 aircraft').

\subsubsection*{Property Claim 8: The probabilistic predictor quantifies uncertainty with sufficient accuracy and fidelity}

The probabilistic trajectory can draw a sample from $p(\boldsymbol{\alpha}, \boldsymbol{\beta})$ when an aircraft is instantiated and then generate a trajectory by solving \eqref{eq:bada_rocd} numerically using \eqref{eq:fpca_drag} and \eqref{eq:fpca_cas} to represent drag and CAS. By drawing an ensemble of samples for a given prediction, it can generate a set of trajectories which can be used to quantify the uncertainty of its mean-mode predictions. This is a critical requirement of the EASA Guidance (Anticipated MOC EXP-07).

As with mean-mode, we divide our claims into different kinds of maneuvers, such as descents, climbs and cruising aircraft, which would in turn be sub-divided into aircraft types (not pictured for brevity). Taking the example of a B738 descending aircraft, the task is to provide evidence that demonstrates it quantifies uncertainty with sufficient accuracy and fidelity.

For measuring fidelity and following the W-process for assuring training (e.g. P8.1.1 for descending B738s), we need to consider how well relevant quantities match the real-world. However, we are comparing a generated distribution of samples to the real-world distribution rather than comparing averages. Thus we need to provide evidence on whether the statistical distance between these distributions is reasonable (see `deep dive' below).

\begin{tcolorbox}[colback=white, colframe=gray!50!black, breakable]

\textbf{Deep dive on Property Claim 8.1.1\\
The probabilistic predictor captures relevant distributions for descending Boeing 738s}\\

Figure \ref{fig:b738_prob} shows the descent profile (top panels) and true airspeed (bottom panels) for a sample of descending Boeing 737-800s (B738s). This compares the real-world data (left panel, blue lines) with the data generated by the probabilistic predictor (right panel, red lines). Because the BADA model is deterministic, it only produces a single prediction for an aircraft descending from a given flight level (green lines). 

\begin{center}
\includegraphics[
    width=0.8\textwidth
]{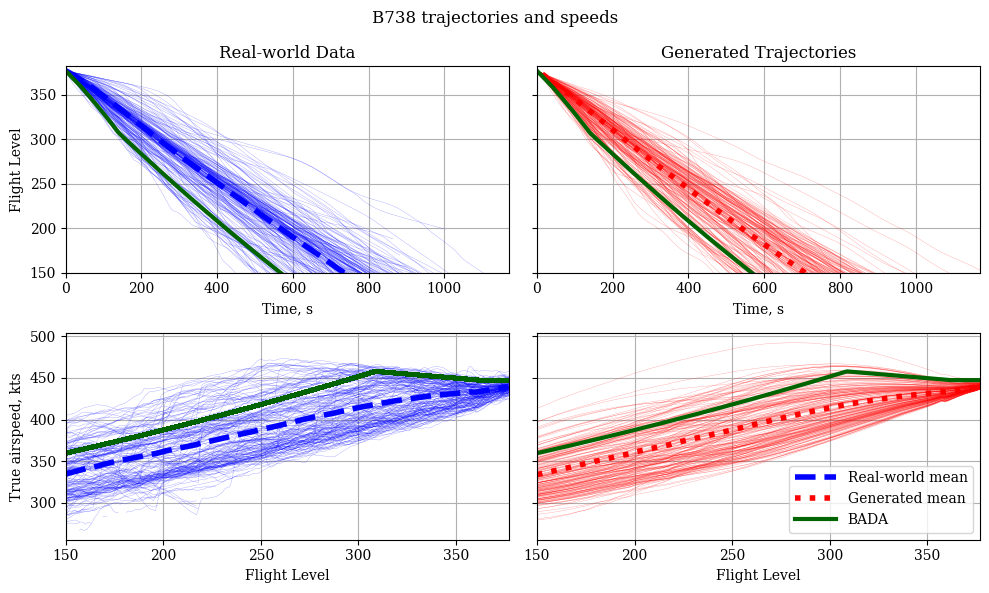}
\captionof{figure}{Figure originally from \citet{DTpaper}. Data (blue lines, left panels) for B738 descending from flight level 360 as compared to BADA (green lines) and trajectories generated by the probabilistic predictor (red lines, right panels). This shows a visual match to descent profile (top panels) and true airspeed (bottom panels).}
\label{fig:b738_prob}
\end{center}

\vspace{1em} 

To provide evidence on how well the probabilistic predictor captures the distribution of calibrated airspeeds (CAS), we determine their statistical distance (E8.1.1a). We assume that CAS is a relevant quantity (A8.1.1a) because it is a directly learnt parameter and serves as a proxy for distance travelled (J8.1.1a). We assume Kolmogorov-Smirnov (KS) and Wasserstein distance are appropriate (A8.1.1b) because they are complementary: whereas KS distance assesses maximum difference in the cumulative distribution function and is dimensionless, Wasserstein distance quantifies the cost required to transform one distribution into the other, has more sensitivity to shifts in the tails and is related to the unit of interest (J8.1.1b). We assume there are no fixed thresholds for failure (A8.1.1c). This is because unlike the mean-mode, BADA cannot be used as a benchmark. Determining a threshold for these measures is likely to be highly dependent on the data distribution for a given aircraft, how many samples are available and its modality (J8.1.1c). As a starting point, we are guided by whether distributions appear visually similar and use improvements on our own model as a measure. Similarly to CAS, we generate Evidence for other relevant quantities, namely rate of descent (E8.1.2) and time to bottom of descent (E8.1.3). Each of these could have additional Assumptions and Justifications. For example, time to bottom of descent is a relevant quantity for ATC as TP methods are used within operational systems to predict future level occupancy to inform ATCO decision making (not pictured).

Table \ref{table:b738_prob} shows the results of these statistical tests for each of the quantities and statistical distances. Note that the rate of descent and calibrated airspeed data is separated by whether the aircraft was above or below the transition point (i.e.\ the altitude where the aircraft's true airspeed is equal to its Mach number). For a complete set of evidence for different descending aircraft types, see \citet{hodgkin2025probabilisticsimulationaircraftdescent}.

\begin{center}
\captionof{table}{Comparison of generated trajectories all descending B738s against the held out test set. * Transition refers to whether the data is above or below the transition point.}
\begin{tabular}{ccccc}
    \hline
    Measure & Transition* & KS distance  & Wasserstein distance \\
    \hline
    Calibrated Airspeed (knots) & Above & 0.159  & 2.73 \\
    Calibrated Airspeed (knots) & Below & 0.149 & 4.49 \\
    Rate of Descent (feet/min) & Above & 0.145 & 180 \\
    Rate of Descent (feet/min) & Below & 0.149 & 187 \\
    Time to Bottom of Descent (s) & - & 0.158 & 31.8 \\
    
\end{tabular}
\label{table:b738_prob}
\end{center}

\end{tcolorbox}

To measure accuracy and follow the W-process for inference verification (P8.1.2), we again consider the vertical, cross-track and along-track dimensions along with relevant quantities like CAS over the same look-ahead time as for mean-mode in P7 (i.e.\ every 6 seconds and from clearance to clearance). However, we need to compare the ensemble of predictions for a given trajectory to a single ground truth value, and so cannot use the MAE. Instead, we consider two complementary Evidence elements:
\begin{itemize}
    \item \textbf{Continuous Ranked Probability Score}: this is a generalisation of the MAE, which treats the ground truth value as a step distribution and considers the area between it and the ensemble's distribution. Although the values of a single CRPS are difficult to interpret, it can be used as a relative measure which allows us to assess how well one aspect of the predictor (e.g.\ descents) captures uncertainty versus another (e.g.\ climbs), such as by calculating the CRPS skilfulness score.
    \item \textbf{Calibration Curve}: this is essentially a probability-probability (PP) plot that assesses how well the average of predictions matches the evidence \cite{thelen_comprehensive_2023}. A perfectly calibrated model would form a straight line at a 45 degree angle. We seek predicted values at the 2.5th and 97.5th quantiles that contain the data 95\% of the time whilst using P3.1.1. as a constraint on the sharpness of the confidence bounds.
\end{itemize}

\subsection*{Strategy 4: Argument over accuracy and fidelity of AI agents via the Digital Twin}

\begin{figure}
\centering
\includegraphics[width=1\textwidth]{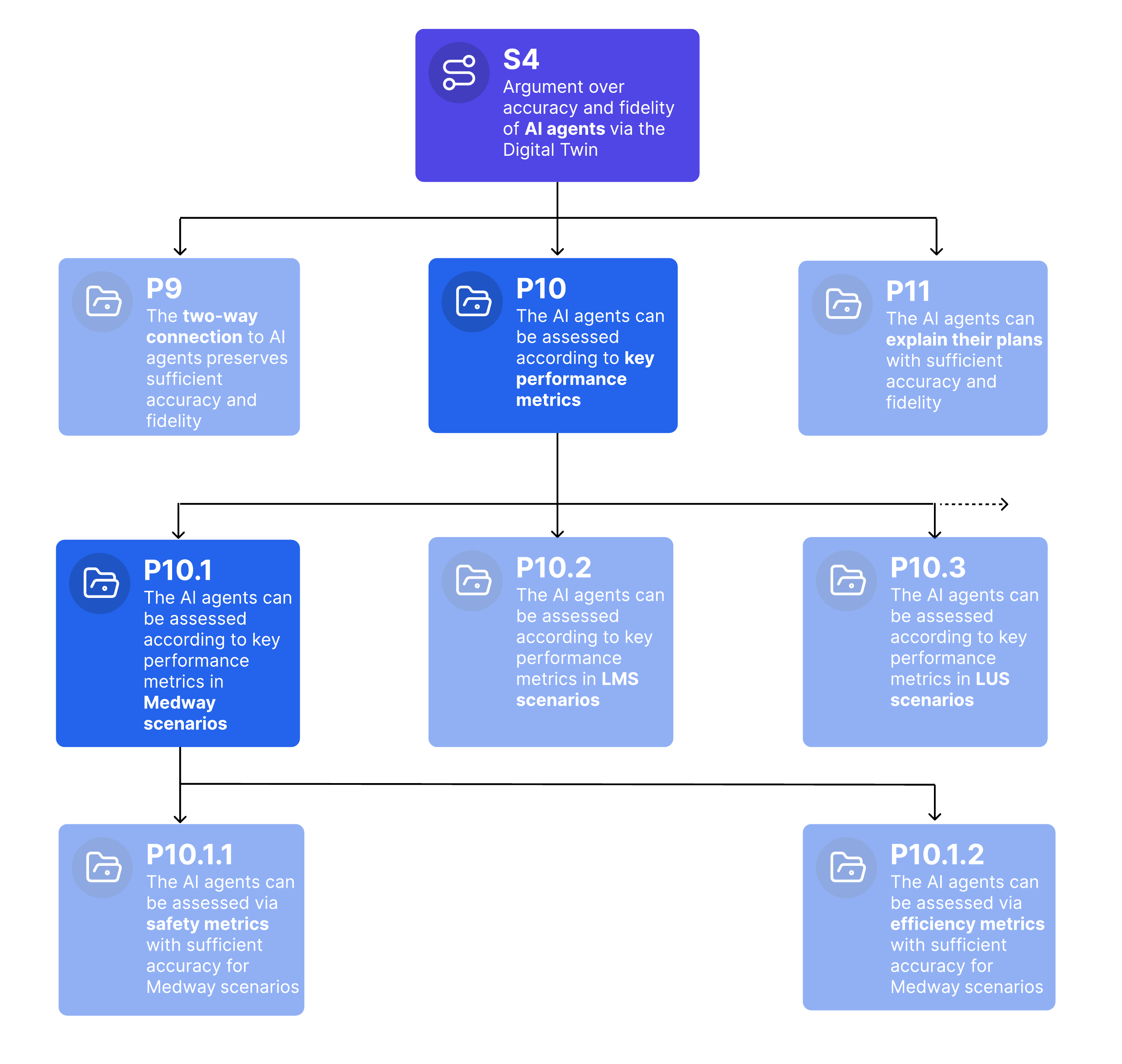}
\caption{Abridged argument over interoperability with AI agents. S=Strategy, P=Property Claim.}
\label{fig:strategy-agents}
\end{figure}

Figure \ref{fig:strategy-agents} shows selected elements from the argument over AI agent enablement. We provide additional context and then elaborate upon the top-level Property Claims.

\subsubsection*{Additional context}

The Digital Twin provides a simulation environment in which AI agents can be trained and evaluated on real-world traffic samples, where a sector of airspace for a given day can be simulated with the rest following replay mode. It is designed to be agent-agnostic with agents only needing to connect to the Digital Twin via a REST API and pass a structured set of JSONs which follow a particular format for issuing clearances. In return, the Digital Twin passes current information on aircraft positions based on either replay or trajectory projections. There is also a suite of gymnasium environments supporting multi-agent scenarios and specification of rewards, actions and state space.

There are variety of AI agents which have been developed using the Digital Twin as a training and testing environment. These include:
\begin{itemize}
\item \textbf{Reinforcement Learning agent}: This leverages behavioural cloning and proximal policy optimisation \cite{RL_paper}.
\item \textbf{Rules-based agent}: One version solves two-aircraft interactions based on the typical strategies an ATCO would use whilst another is based upon a restricted set of pathways aircraft may travel along \cite{rule_based_agent, bluebird_mallard}.
\item \textbf{Optimisation agent}: This treats the problem as one which maximises efficiency constrained by safety considerations.
\end{itemize}

Quantitative metrics can be computed at runtime, including safety indicators and efficiency metrics. These metrics assist with assessing the performance of an agent. However, these metrics quantify specific aspects of the controlling task and are insufficient to gain insights to overall proficiency, where controllers must balance safety, efficiency, and orderliness in a highly context-specific manner. To that end, the Digital Twin also contains a Human Machine Interface (HMI) which visualises the airspace configuration, aircraft intention and agent plans in a manner similar to ATCO operations. This enables human-in-the-loop evaluation of AI agent assessment, which has been used to assess how well AI perform in a modified version of the Basic Training course that prospective ATCOs must undergo in the UK\footnote{Some competencies that trainee ATCOs must demonstrate, such as radio-telephony, are ruled out of scope in the agent validation framework. The interested reader is referred to \citet{Agent_validation_framework} for a detailed discussion on how the Basic Training course is modified for AI agent assessment within this framework.}. For details on the outcome of these evaluations the interested reader is referred to \citet{Agent_validation_framework}.

\subsubsection*{Property Claim 9: The two-way connection to the AI agents preserves sufficient accuracy and fidelity}
The performance of the AI agents depends on the accuracy, fidelity, and reliability of the bidirectional connection between the agents and the Digital Twin. All information relevant to agent decision making, including radar data, coordination information and any agent-issued commands, must be transmitted completely and correctly. Each message must be associated with the correct simulation timestamp, and safeguarded against partial failures such as dropped or duplicated packets.

Connection latency must be low enough for the agents to operate within the required control timescales, ideally within a single radar update cycle (approximately 6 seconds). Meeting this requirement necessitates validation of both message integrity and end-to-end timing performance.

For LAN deployments, validation should include sustained throughput measurements and round-trip latency testing under representative load conditions. Integrity checks such as hashing and checksums verify that messages remain unaltered in transit.

Assurance of the REST API requires well-defined schemas, strong authentication and authorisation mechanisms, and secure code practices supported by static analysis. Runtime assurance through dynamic testing, contract testing and performance testing verifies correct behaviour under load, to ensure secure and reliable operation.

\subsubsection*{Property Claim 10: The AI agents can be assessed according to key performance metrics}

An AI agent must be able to pass key performance metrics that human ATCOs also need to achieve. It must do this in both the ATC training sector of Medway (P10.1) but also each bandboxed sector of UK airspace, such as London Middle Sector (P10.2) and London Upper Sector (P10.3). For a given sector, we need to ensure that a representative number of samples is used over different dates and times, giving a reasonable diversity of scenarios (e.g.\ including seasonality and traffic patterns). For Medway, this means each of the 12 assessment versions that ATCO trainees can be administered. We also need to account for contingencies such as storms; such unusual scenarios may need to be generated via the synthetic scenario generation algorithm described in P5.

The Digital Twin enables an AI agent to be tested during a simulation and provide real-time information and logs of their performance. There are a number of metrics it calculates to assess safety (e.g. P10.1.1 for Medway) and efficiency (e.g. P10.1.2 for Medway). These include:
\begin{itemize}
    \item \textbf{Loss of separation}: Measures whether separation standards for en route airspace have been maintained. In en route airspace the separation standard is typically 5 nautical miles laterally or 1,000~ft vertically \cite{CAA_CAP493}. 
    \item \textbf{Technical safety}: Measures whether a fail-safe method of operation has been employed by the agent. It expands traditional conflict detection rollouts with significant uncertainty bounds and modifications designed to prove that the separation standards between aircraft are guaranteed to be achieved with no further intervention.
    \item \textbf{Coordination compliance}: Coordinations are agreements with neighboring sectors about what location a flight should be passed between them. This checks whether the aircraft leaves the sector at the correct flight level and sufficiently near to the correct fix.
    \item \textbf{Airspace excursions}: An aircraft under AI control should not leave the sector's defined geographic boundaries prior to reaching its designated exit point, unless a revised coordination has been explicitly agreed upon. This metric can be quantified as the total time during which controlled aircraft, across a scenario, operate outside the sector earlier than authorised.
    \item \textbf{Fuel burn}: This is a measure of efficiency which assesses how much fuel an aircraft burns when traversing the sector, which can be a proxy for understanding aircraft performance and issuing appropriate vectors and timely climbs and descents.
    \item \textbf{Traffic complexity}: Typical traffic load volumes for a given time period for a sector and how they compare to the agent-controlled volume can give an indication as to how well the agent is managing airspace complexity.
\end{itemize}

Maintaining separation standards at all times is vital and an AI agent would be expected to satisfy this metric at all times. For other measures, the threshold should be determined by whether they can achieve parity with the performance of experienced ATCOs according to hypothesis testing. The weighting between certain metrics could be determined through a combination of consultation with air traffic controllers and semi-supervised learning of these thresholds based on how they correlate to human-assigned Basic Training scores.

The calculation of fuel burn, whilst useful, is complicated by the unavailability of this data; instead we must rely on estimates derived from the BADA model. Simpler measures such as transit time could be used in lieu but the performance of a given aircraft needs to be factored into such calculations.

To validate a given AI agent, data needs to be divided into a training and testing set. This is especially critical for agents whose training may not extrapolate well. For Basic Training, this is naturally divided into the formative scenarios which are used to train human air traffic controllers and the summative scenarios used to assess them, respectively. For real-world scenarios this may involve having certain sectors of UK airspace or certain time intervals held out.

\subsubsection*{Property Claim 11: The AI agents can explain their plans with sufficient accuracy and fidelity}

For ATCOs to assess whether aircraft are being controlled safely and efficiently, the HMI should capture all salient components and information for an expert to understand the context in which an agent is issuing an instruction. In other words, it should reflect what they are used to seeing in operations. Existing regulations that set the standards for the HMI used by ATCOs in operational systems can be used as the basis for suitability of the Digital Twin's HMI (see e.g. CAP 670 \cite{CAA_CAP670}). This is also recognised in `Human factors' for Level 1A AI in EASA Guidance, which states that existing guidelines and requirements for interface design should be used. Therefore, details such as the sector boundaries, aircraft selected flight levels and current flight levels, are depicted in a manner ATCOs are familiar with. This would also include text-to-voice software to emulate radio transmissions with consistent and appropriate nomenclature. The data for a simulation can be logged, providing the means to review the behaviour of the AI agent. In agreement with EASA Guidance, these logs should be completed over the course of a simulation (Anticipated MOC EXP-09-1), contain all relevant information to identify deviations (Anticipated MOC EXP-09-2) and are easily retrievable for those who need it (Anticipated MOC EXP-09-3). To that end, the radar and clearance logs, along with the safety and efficiency metrics, can be recorded and updated in real-time.

\begin{figure}
\centering
\includegraphics[width=1\textwidth]{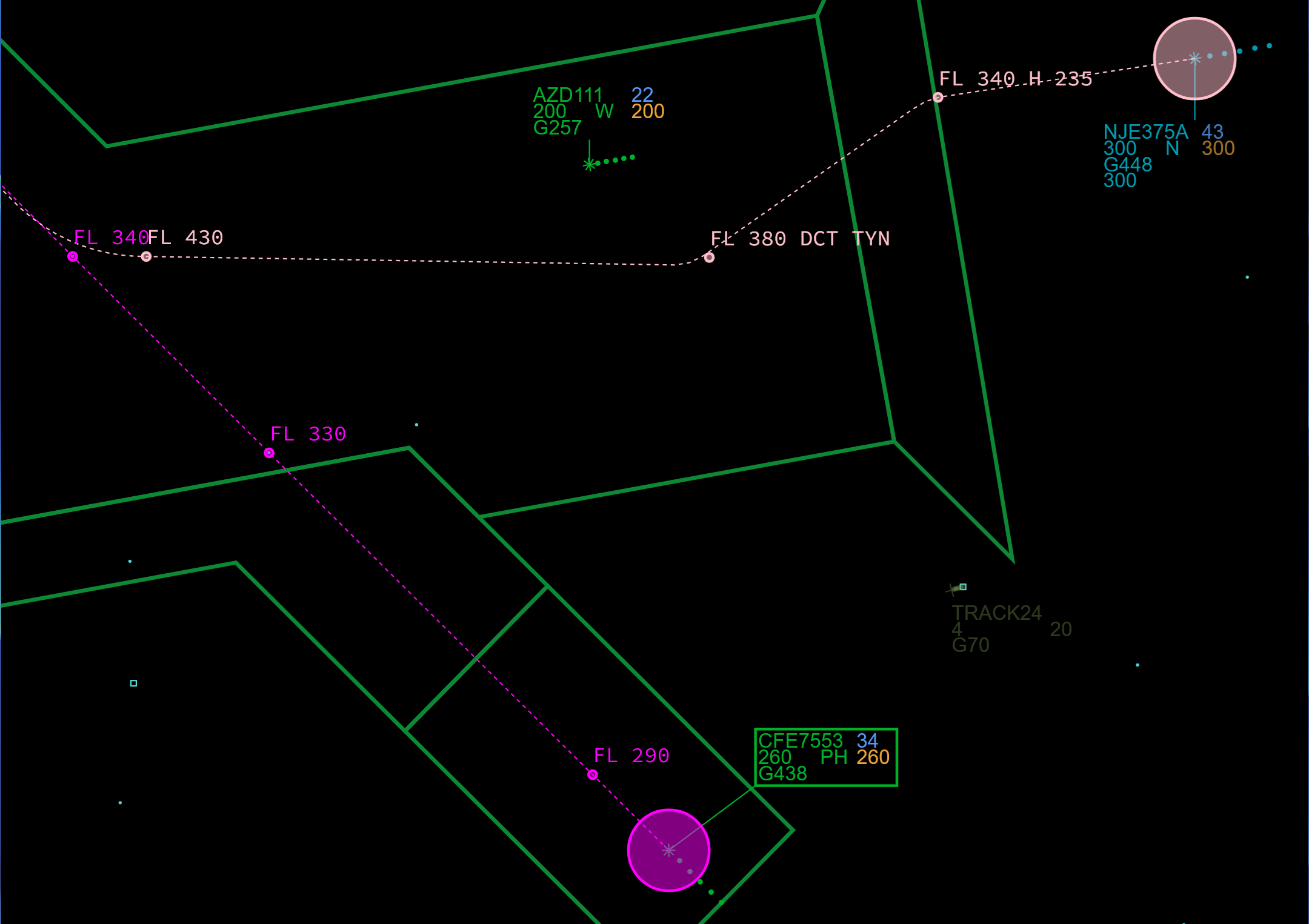}
\caption{Figure originally from \citet{rule_based_agent}. Visualisation of the optimisation agent solving an exercise in Basic Training. Aircraft relevant to the decision-making of the agent with respect to the subject aircraft (CFE7553, magenta) are highlighted with a coloured circle (NJE375A, pink). The agent's plan for the subject aircraft and all aircraft jointly optimised along with it are also shown in their respective colours. The green outline depicts the sector boundaries.}
\label{fig:optimisation_plans}
\end{figure}

A key competency in Basic Training is planning, so AI agents need to show their plans. This is likely to be agent-dependent with a separate plan for each agent. Humans use hand-written strips to record key details of each flight such as their selected flight level, and note their plan for them. They are also able to explain their plans verbally. In order for humans to assess the plans the AI is making, the Digital Twin can use the AI's plans to emulate these strips. They are also depicted in the HMI in a unique colour, with lines and points for the planned trajectory and clearances, respectively (Fig \ref{fig:optimisation_plans}). The horizon over which a plan is depicted may depend on the agent: for example some rules-based agents may only plan for the next interaction and instruction whereas a search-based agent might devise plans along the entire flight through the sector. The relevant aircraft which informed an agent's decision may be highlighted with circles (Fig \ref{fig:optimisation_plans}). What constitutes a relevant aircraft will depend on the kind of agent and needs to be assured on an agent-by-agent basis. In either case, experts can assess whether these were what they would regard as relevant aircraft. This approach is in concordance with EASA Guidance (Anticipated MOC EXP-12), which details the need for information about goals (e.g. the coordinations), the context for the decision making (e.g. relevant aircraft), information on the usual way of reasoning (e.g. strips), trade-offs (e.g components of a reward function) and sources used for the decision.  Another potentially agent-agnostic approach is the use of counterfactual explanations, which indicate the smallest change in conditions that would result in a different decision  (e.g. `the agent vectored aircraft A right because of a loss of separation which would occur with aircraft B; if aircraft B were 10 flight levels higher, it would not have been issued an action.') These have been utilised as an explanation for black box models \cite{counterfactuals_2018}.

Although agents may be able to provide a large amount of detail, it is essential to present explanations at a level of abstraction appropriate to the assessor's expertise and the real-time constraints of assessments (Anticipated MOC EXP-13). It is not helpful or practical for an ATCO assessor viewing actions in real-time to see every branch of an agent's internal decision tree. However, it is helpful to see a clear visual indication of which aircraft influenced a given decision. The validity of these explanations is a necessary objective of the EASA Guidance (Objective EXP-17), which naturally comes from human-in-the-loop assessments of planning competencies.

\newpage

\section{Discussion and conclusion}
\label{sec:discussion}

This paper utilised the Trustworthy and Ethical Assurance (TEA) framework to develop an assurance case for a probabilistic Digital Twin of operational airspace. The assurance case develops an emerging community of practice and connects novel technologies developed as part of the Digital Twin, such as probabilistic trajectory generation and LLM-driven scenario generation, to the emerging regulatory guidance in ATC.

We constrained the scope of our ongoing assurance case by a) specifying important contextual information about the probabilistic Digital Twin that is intended to be used as a training and testing environment for AI agents ; and b) operationalising key terms such as `sufficient accuracy and fidelity' through concrete Property Claims. These claims were grouped together based on four strategies focusing on the accuracy and fidelity of interconnected components of the Digital Twin, namely the data pipeline, its virtual representation of this data, trajectory prediction, and its interoperability with AI agents. The structure of each of these strategies was informed by both established practice and publications, as well as the emerging literature and guidance around AI and Digital Twins both generally and in ATM. We provided additional detail on two Property Claims through evidenced deep dives on 1) the ability of GPT-5 to accurately translate prompts into  synthetic airspace scenarios; and 2) the ability of the probabilistic predictor to accurately generate trajectories for descending B738s. Both deep dives included the Assumptions and Justifications for the choice of Evidence.

Probabilistic Digital Twins provide many challenges to developers, regulators, auditors, and other stakeholder groups, due to the dynamic use of data, their predictive capabilities, and their inherent uncertainty. Ensuring seamless interoperability with AI agents adds a further layer of complexity. This work provides an empirically grounded use case for how to assure their accuracy and fidelity within an intended environment. It makes transparent assumptions and justifications that are often only implicit in published research. The focus here is on accuracy and fidelity, but it could form a component of a broader case to help address goals such as safety and security, fairness, in line with the CAA's strategy for AI \cite{caa_cap2970}. For operational usage of Digital Twins, safety assurance would be paramount, and entail more stringent thresholds for accuracy and fidelity as well as a range of additional evidence or evaluations, such as searching for and characterising failure modes, or using the Value at Risk (value guaranteed not to exceed this amount for a given quantile of a distribution) instead of the mean, utilising risk tables from EASA's detailed specifications, and performing safety envelope testing \cite{safety_envelope}. Due to the modularity of argument-based assurance, our existing assurance case could be extended to meet this need.

Our Digital Twin is AI-enabled, which raises an additional question: can the Digital Twin be assured separately to each AI agent? We have assumed that this is the case, but operationalising a given AI agent and the Digital Twin as a single system may require a separate assurance case for each. For example, there could be emergent behaviour arising from their interaction which cannot be anticipated from assuring the individual components. Nor is it clear how to completely separate the veracity of synthetic data from the agents that are trained upon it. Similar questions would arise for interoperating with airport or terminal control Digital Twins.

The expected incorporation of live data assimilation \cite{Assimilation_paper} would require online validation as an element of the assurance case; in this case, the validation is essentially a lagging indicator. Indeed, EASA Guidance and the FAA roadmap \cite{faa_roadmap} note that this kind of real-time learning is incompatible with current certification processes. One approach is to measure the current performance and the corresponding model uncertainty in previous time instants and extrapolate to future time instants; however, this then requires assurance of the extrapolation method itself \cite{thelen_comprehensive_2023}.

We plan to conduct a workshop in March 2026, which surveys opinion and feedback from stakeholders, such as key regulators, governmental bodies, and experts in Digital Twin validation. This will delineate factors for the assurance of Digital Twins that are both generic and ATM-specific, thereby making our research more robust and generalisable. As well as strengthening the existing case, it may help `fill in the blanks' for some of the aforementioned issues around live data assimilation and AI agent assurance. Finally, this work will also feed back into the TEA platform, to improve its functionality for the benefit of the wider community and to display an interactive and iterative version of our assurance case, as well as a compendium of appropriate methods for a given type of case (e.g.\ accuracy).

In conclusion, we have developed an assurance case about the accuracy and fidelity of an AI-enabled Digital Twin of en route UK airspace. It attempts to account for emerging and established guidance around Digital Twins in ATM, while also offering scope for generalisation to other sectors. The probabilistic Digital Twin developed by Project Bluebird has been designed as a training and testing environment for AI agents, but is flexible and extensible enough to form a template for the many other potential applications, and interface with other Digital Twins. Finally, it helps contribute to an emerging community of practice around the use and assurance of Digital Twins.

\bibliography{references}

\end{document}